%% file: 2022-fairness-framework.tex
\documentclass[sigconf,natbib,screen=true]{acmart}
\input{packages}
\input{definitions}

\input{meta}
\input{authors}

\title[Probabilistic Permutation Graph Search]{Probabilistic Permutation Graph Search: \\Black-Box Optimization for Fairness in Ranking}

\allowdisplaybreaks

\begin{document}


\begin{abstract}
	There are several measures for fairness in ranking, based on different underlying assumptions and perspectives.
	\ac{PL} optimization with the REINFORCE algorithm can be used for optimizing black-box objective functions over permutations. In particular, it can be used for optimizing fairness measures.
	However, though effective for queries with a moderate number of repeating sessions, \ac{PL} optimization has room for improvement for queries with a small number of repeating sessions.
	
	In this paper, we present a novel way of representing permutation distributions, based on the notion of permutation graphs.
	Similar to~\ac{PL}, our distribution representation, called~\ac{PPG}, can be used for black-box optimization of fairness.
	Different from~\ac{PL}, where pointwise logits are used as the distribution parameters, in~\ac{PPG} pairwise inversion probabilities together with a reference permutation construct the distribution.
	As such, the reference permutation can be set to the best sampled permutation regarding the objective function, making~\ac{PPG} suitable for both deterministic and stochastic rankings.
	Our experiments show that~\ac{PPG}, while comparable to~\ac{PL} for larger session repetitions (i.e., stochastic ranking), improves over~\ac{PL} for optimizing fairness metrics for queries with one session (i.e., deterministic ranking).
	{Additionally, when accurate utility estimations are available, e.g., in tabular models, the performance of \ac{PPG} in fairness optimization is significantly boosted compared to lower quality utility estimations from a learning to rank model, leading to a large performance gap with PL.}
	Finally, the pairwise probabilities make it possible to impose pairwise constraints such as ``item $d_1$ should always be ranked higher than item $d_2$.''
	Such constraints can be used to simultaneously optimize the fairness metric and control another objective such as ranking performance.
\end{abstract}



\maketitle

\acresetall

\input{sections/01-introduction}
\input{sections/02-related}
\input{sections/03-method}
\input{sections/04-experiments}
\input{sections/05-results}
\input{sections/06-conclusion}

\section*{Code and data}
To facilitate the reproducibility of our results, this work only made use of publicly available data and our experimental implementation is publicly available at \url{https://github.com/AliVard/PPG}.

\begin{acks}
This research was supported by Elsevier, the Netherlands Organisation for Scientific Research (NWO)
under pro\-ject nr 612.\-001.\-551, Ahold Delhaize, and the Hybrid Intelligence Center, a 10-year program funded by the Dutch Ministry of Education, Culture and Science through the Netherlands Organisation for Scientific Research, \url{https://hybrid-intelligence-centre.nl}.
All content represents the opinion of the authors, which is not necessarily shared or endorsed by their respective employers and/or sponsors.
\end{acks}

\appendix
\input{sections/0A-appendix}

\input{sections/0B-table}

\bibliographystyle{ACM-Reference-Format}
\balance
\bibliography{references}

\end{document}

%% file: packages.tex
\PassOptionsToPackage{dvipsnames}{xcolor}

\usepackage{dsfont}
\usepackage{acronym}
\usepackage[noend]{algorithmic}
\usepackage[ruled,vlined,linesnumbered]{algorithm2e}
\usepackage{bbm}
\usepackage{beramono}
\usepackage{bm}
\usepackage{booktabs}
\usepackage[skip=0pt]{caption}
\usepackage{cleveref}
\usepackage[T1]{fontenc}

\usepackage{graphicx}
\usepackage{hyperref}
\usepackage{listings}
\usepackage{multirow}
\usepackage{subcaption}
\usepackage{tabularx}
\usepackage[dvipsnames]{xcolor}
\usepackage[inline]{enumitem}
\usepackage{numprint}
\usepackage[normalem]{ulem}

\usepackage{balance}

\usepackage{mathrsfs}
\usepackage{mathtools}
\usepackage{xspace}

\usepackage{tikz}
\usetikzlibrary{positioning, fadings,patterns,fit}
\usetikzlibrary{arrows}

\makeatletter
\newif\ifAC@uppercase@first%
\def\Aclp#1{\AC@uppercase@firsttrue\aclp{#1}\AC@uppercase@firstfalse}%
\def\AC@aclp#1{%
  \ifcsname fn@#1@PL\endcsname%
    \ifAC@uppercase@first%
      \expandafter\expandafter\expandafter\MakeUppercase\csname fn@#1@PL\endcsname%
    \else%
      \csname fn@#1@PL\endcsname%
    \fi%
  \else%
    \AC@acl{#1}s%
  \fi%
}%
\def\Acp#1{\AC@uppercase@firsttrue\acp{#1}\AC@uppercase@firstfalse}%
\def\AC@acp#1{%
  \ifcsname fn@#1@PL\endcsname%
    \ifAC@uppercase@first%
      \expandafter\expandafter\expandafter\MakeUppercase\csname fn@#1@PL\endcsname%
    \else%
      \csname fn@#1@PL\endcsname%
    \fi%
  \else%
    \AC@ac{#1}s%
  \fi%
}%
\def\Acfp#1{\AC@uppercase@firsttrue\acfp{#1}\AC@uppercase@firstfalse}%
\def\AC@acfp#1{%
  \ifcsname fn@#1@PL\endcsname%
    \ifAC@uppercase@first%
      \expandafter\expandafter\expandafter\MakeUppercase\csname fn@#1@PL\endcsname%
    \else%
      \csname fn@#1@PL\endcsname%
    \fi%
  \else%
    \AC@acf{#1}s%
  \fi%
}%
\def\Acsp#1{\AC@uppercase@firsttrue\acsp{#1}\AC@uppercase@firstfalse}%
\def\AC@acsp#1{%
  \ifcsname fn@#1@PL\endcsname%
    \ifAC@uppercase@first%
      \expandafter\expandafter\expandafter\MakeUppercase\csname fn@#1@PL\endcsname%
    \else%
      \csname fn@#1@PL\endcsname%
    \fi%
  \else%
    \AC@acs{#1}s%
  \fi%
}%
\edef\AC@uppercase@write{\string\ifAC@uppercase@first\string\expandafter\string\MakeUppercase\string\fi\space}%
\def\AC@acrodef#1[#2]#3{%
  \@bsphack%
  \protected@write\@auxout{}{%
    \string\newacro{#1}[#2]{\AC@uppercase@write #3}%
  }\@esphack%
}%
\def\Acl#1{\AC@uppercase@firsttrue\acl{#1}\AC@uppercase@firstfalse}
\def\Acf#1{\AC@uppercase@firsttrue\acf{#1}\AC@uppercase@firstfalse}
\def\Acfi#1{\AC@uppercase@firsttrue\acfi{#1}\AC@uppercase@firstfalse}
\def\Ac#1{\AC@uppercase@firsttrue\ac{#1}\AC@uppercase@firstfalse}
\def\Acs#1{\AC@uppercase@firsttrue\acs{#1}\AC@uppercase@firstfalse}
\robustify\Aclp
\robustify\Acfp
\robustify\Acp
\robustify\Acsp
\robustify\Acl
\robustify\Acf
\robustify\Acfi
\robustify\Ac
\robustify\Acs
\makeatother

\def\AC@@acro#1[#2]#3{%
  \ifAC@nolist%
  \else%
  \ifAC@printonlyused%
    \expandafter\ifx\csname acused@#1\endcsname\AC@used%
       \item[\protect\AC@hypertarget{#1}{\acsfont{#2}}] #3%
          \ifAC@withpage%
            \expandafter\ifx\csname r@acro:#1\endcsname\relax%
               \PackageInfo{acronym}{%
                 Acronym #1 used in text but not spelled out in
                 full in text}%
            \else%
               \dotfill\pageref{acro:#1}%
            \fi\\%
          \fi%
    \fi%
 \else%
    \item[\protect\AC@hypertarget{#1}{\acsfont{#2}}] #3%
 \fi%
 \fi%
 \begingroup
    \def\acroextra##1{}%
    \@bsphack
    \protected@write\@auxout{}%
       {\string\newacro{#1}[\string\AC@hyperlink{#1}{#2}]{\AC@uppercase@write #3}}%
    \@esphack
  \endgroup}
  

%% file: definitions.tex
\definecolor{ali}{RGB}{0, 150, 0}
\definecolor{mdr}{RGB}{255, 127, 0}

\acrodef{PPG}{probabilistic permutation graph}
\newcommand{\ourmethod}{\ac{PPG} search\xspace}
\newcommand{\ourmethodtitle}{Probabilistic Permutation Graph Search}

\newcommand{\myparagraph}[1]{\emph{\bfseries #1.}}

\newcommand{\mExpect}[2]{\ensuremath{\mathbb{E}_{#1}\left[#2\right]}}

\newcommand{\permcond}{\rho}
\newcommand{\permspace}{\ensuremath{\mathcal{P}}\xspace}

\acrodef{IR}{information retrieval}
\acrodef{EM}{Expectation-Maximization}
\acrodef{LTR}{learning to rank}
\acrodef{SERP}{Search Engine Results Page}
\acrodef{CLTR}{counterfactual learning to rank}
\acrodef{OLTR}{Online Learning to Rank}
\acrodef{IPS}{inverse propensity scoring}
\acrodef{PBM}{Position-Based Model}
\acrodef{CM}{Cascade Model}
\acrodef{ARP}{Average Relevance Position}
\acrodef{DCG}{Discounted Cumulative Gain}
\acrodef{DLA}{Dual Learning Algorithm}
\acrodef{MNAR}{Missing-Not-At-Random}
\acrodef{mPMM}{modified Poisson Mixture Model}
\acrodef{DCM}{Dependent Click Model}
\acrodef{UBM}{User Browsing Model}
\acrodef{CCM}{Click Chain Model}
\acrodef{CTR}{click-through rate}
\acrodef{AC}{affine correction}
\acrodef{i.i.d.}{independent and identically distributed}
\acrodef{CLT}{Central Limit Theorem}
\acrodef{NDCG}{Normalized Discounted Cumulative Gain}
\acrodef{MAP}{Mean Average Precision}
\acrodef{DTR}{disparate treatment ratio}
\acrodef{MSE}{mean-squared error}
\acrodef{PL}{Plackett-Luce}
\acrodef{EEL}{expected exposure loss}
\acrodef{RL}{reinforcement learning}
\acrodef{MC}{Monte-Carlo}

\theoremstyle{definition}

\theoremstyle{definition}

\newtheorem{example}{Example}[]

\allowdisplaybreaks

\makeatletter
\renewcommand\paragraph{\@startsection{paragraph}{4}{\parindent}%
  {-.1\baselineskip \@plus -1\p@ \@minus -.1\p@}%
  {-2\p@}%
  {\ACM@NRadjust{\@parfont\@adddotafter}}}
\makeatother  

\newcommand{\ROne}[1]{\textcolor{teal}{\textbf{[R1]}}}
\newcommand{\RTwo}[1]{\textcolor{blue}{\textbf{[R2]}}}
\newcommand{\RThree}[1]{\textcolor{cyan}{\textbf{[R3]}}}

%% file: meta.tex
\copyrightyear{2022} 
\acmYear{2022} 
\setcopyright{acmlicensed}
\acmConference[SIGIR '22]{Proceedings of the 45th International ACM SIGIR Conference on Research and Development in Information Retrieval}{July 11--15, 2022}{Madrid, Spain}
\acmBooktitle{Proceedings of the 45th International ACM SIGIR Conference on Research and Development in Information Retrieval (SIGIR '22), July 11--15, 2022, Madrid, Spain}
\acmPrice{15.00}
\acmDOI{10.1145/3477495.3532045}
\acmISBN{978-1-4503-8732-3/22/07}

%% file: authors.tex
\author{Ali Vardasbi}
\affiliation{%
  \institution{University of Amsterdam}
  \city{Amsterdam}
  \country{The Netherlands}
}
\email{a.vardasbi@uva.nl}

\author{Fatemeh Sarvi}
\affiliation{%
  \institution{AIRLab, University of Amsterdam}
  \city{Amsterdam}
  \country{The Netherlands}
}
\email{f.sarvi@uva.nl}

\author{Maarten de Rijke}
\affiliation{%
  \institution{University of Amsterdam}
  \city{Amsterdam}
  \country{The Netherlands}
}
\email{m.derijke@uva.nl}

%% file: sections/01-introduction.tex

\section{Introduction}
\label{sec:intro} 
Several fairness measures are being considered in the search and recommendation literature.
Different measures have been proposed based on different definitions of fairness, aimed at different environments.
For instance,~\citet{singh2018fairness} consider the \ac{DTR}, which ensures equity of exposure: each group should get exposed proportional to their utility.
\Ac{DTR} for group fairness can be used for deterministic rankers, which produce one ranking per query, as well as stochastic rankers, with different randomly sampled rankings per query.
It also makes no assumptions about the utility values of the ranked items.
In contrast, the~\ac{EEL} is based on the premise that groups (or individuals) with the same relevance level should have equal expected exposures~\citep{diaz2020evaluating}.
By definition,~\ac{EEL} should be used for stochastic rankers.
It is also assumed in~\ac{EEL} that the relevance levels have discrete values.

\myparagraph{Optimizing fairness measures}
The goal of this paper is not to list and compare different fairness definitions; much has already been written about this~\citep{verma2018fairness, raj2020comparing}.
Neither do we want to unify different fairness measures, since they deal with different aspects of fairness, and more importantly, it has been shown that there is an inherent trade-off between some fairness conditions~\citep{kleinberg2017inherent}, which makes it impossible to have a unified fairness measure.
What we aim to accomplish is to provide a general framework that can be used for \emph{optimizing any fairness measure}.

We focus on post-processing methods for fairness and assume that the utility value of the items is given, either from external sources such as unbiased clicks or an estimate computed by a learning to rank~(\acs{LTR}\acused{LTR}) model.
In order to remain general, the framework should work with black-box access to the function that evaluates ranking fairness.
Optimization of fairness in ranking is a special case of permutation optimization, and \ac{RL} is usually the default paradigm for combinatorial optimization: 
A model-free policy-based~\ac{RL} can be used for permutation optimization~\citep{bello2016neural}.
Using the well-known REINFORCE algorithm~\citep{williams1992simple}, and sampling from a~\ac{PL} distribution, it is possible to optimize any fairness objective function on permutations~\citep{singh2019policy}.

A \ac{PL} distribution with the REINFORCE algorithm works very well for optimizing stochastic ranking evaluation metrics such as~\ac{EEL}.
However, there are two directions in which PL-based optimization has room for improvement.
First, when the number of repeating sessions for a query is very small, the high variance of PL~\citep{gadetsky2020low} leads to sub-optimal results.
Second, when an accurate estimate of the utilities is available, the solutions obtained from PL are only slightly improved over the case of noisy utility values.
The first case has practical importance, as it is desirable to have a good fairness solution as soon as possible, before there are a large number of repeating sessions, i.e., a method that provides fair solutions both for deterministic and stochastic rankings.
The second case has both theoretical and practical importance.
From a theoretical point of view, since fairness measures usually depend on utility values,\footnote{Both~\ac{EEL} and~\ac{DTR} metrics do.} noise in utility estimates propagates into the fairness optimization, whereas with accurate utility values, the performance of the fairness optimizer is isolated and not affected by noise from utility values.
This gives a more accurate comparison between different optimization algorithms.
From a practical point of view, given enough historical sessions, unbiased and accurate estimates of utility can be obtained from clicks~\citep{joachims2017unbiased,vardasbi2021mixture, zoghi2016click, pmlr-v115-li20b}.
For instance, extremely high performance rankings can be achieved from tabular models in practice~\citep{10.1145/2911451.2926725}.
We expect to observe a boost in fairness optimization when noisy estimates are replaced with unbiased low noise estimates of utility.

\myparagraph{Black-box permutation optimization}
To bridge the performance gap left by \ac{PL} in the above two cases, we propose a novel permutation distribution for black-box permutation optimization with the REINFORCE algorithm.
Our permutation distribution is based on the notion of permutation graphs~\citep{dushnik1941partially}.
A permutation graph is a graph whose nodes are the items in a ranked list, and whose edges represent inversions in a permutation over the ranked list.
We refer to the initial permutation of the ranked list as the \emph{reference permutation}.
Two examples of permutation graphs are given in Fig.~\ref{fig:permutationgraph}.
In the left graph, over the reference permutation of $[d_1, d_2, d_3, d_4]$, $d_1$ is moved from the first position to the third, causing two inversions $(d_1,d_2)$ and $(d_1,d_3)$.
In the right graph, over the reference permutation of $[d_3, d_2, d_4, d_1]$, $d_1$ is brought forward, and $d_3$ is swapped with $d_2$, causing a total of four inversions, as can be seen in the graph.

\input{sections/figure_tex/01-fig-permutationgraph}

Building on permutation graphs, we define~\acfi{PPG} as a permutation distribution from which the REINFORCE algorithm can sample.
Roughly speaking,~\ac{PPG} is a weighted complete graph, whose edge weights are the probabilities of inverting the two endpoints in a permutation.
Unlike \ac{PL}, in which pointwise logits are used as distribution parameters, \ac{PPG} is constructed from a \emph{reference permutation} together with pairwise inversion probabilities.
The reference permutation in \ac{PPG} helps in deterministic scenarios, as it can be set to the best permutation sampled during the REINFORCE algorithm.
This is not possible for a \ac{PL} distribution, as the deterministic permutation of \ac{PL} comes from sorting the items based on their logits and such a permutation is not necessarily the best sampled permutation during training.

In addition to PPG's gain over PL due to its reference permutation in deterministic as well as accurate relevance estimation scenarios, the pairwise inversion probabilities give PPG another useful property that PL lacks.
In PPG, pairwise constraints can be added during optimization, without any additional computational overhead:
it is enough to set some edge weights to non-trainable parameters.
For example, the constraint of ``item $d_1$ should always be ranked higher than item $d_2$'' can be added by setting to zero the weight of the $(d_1,d_2)$ edge.
Business related, time-aware, or context-aware pairwise constraints can be thought of, all of which are easily handled by \ac{PPG} (see Sec.~\ref{sec:method:constraints}).

Our fairness optimization method is a post-processing method that acts on lists of items, ranked based on utility.
Compared to in-processing methods, such as~\citep{singh2019policy}, a post-processing method allows for richer dynamics, in the sense that having the fairness optimizer separate from the ranker, makes any change on the fairness side independent of the ranking side.
For example, with a post-processing method, if the sensitive groups of attention change over time, either by adding new sensitive features to the existing ones or replacing them, there is no need to re-train the ranker with a new set of group labels.
Our black-box fairness optimization method even adds to this flexibility, as the objective fairness measure itself is allowed to change over time or in different contexts.
Adding to this flexibility, while in-processing fairness optimization methods are an option in feature-based ranking models, they simply cannot be used in tabular models where best rankings are memorized.

\myparagraph{Our contributions}
Our contributions are listed below:
\begin{enumerate}[leftmargin=*,nosep]
    \item We define~\acf{PPG}, a novel permutation distribution.
    \item We present~\ourmethod as a general framework for optimizing fairness in ranking.
    Our method is general in the sense that it works with black-box access to the objective function.
    As such,~\ourmethod can be used in contexts beyond fairness and diversity, to find a permutation that optimizes a general objective function.
    \item We experimentally show the effectiveness of~\ourmethod on two popular objective functions, EEL and DTR, and various public datasets by comparing its performance with state-of-the-art methods for optimizing fairness.
    \item We experimentally verify the effect of pairwise constraints on controlling the ranking performance while optimizing for fairness.
\end{enumerate}

%% file: sections/figure_tex/01-fig-permutationgraph.tex

\begin{figure}
    \begin{tabular}{@{}cc@{}}
        \begin{tikzpicture}
            \node[shape=circle,draw=black] (1) at (0,1) {$d_1$};
            \node[shape=circle,draw=black] (2) at (1,0.5) {$d_2$};
            \node[shape=circle,draw=black] (3) at (0,0) {$d_3$};
            \node[shape=circle,draw=black] (4) at (-1,0.5) {$d_4$};
        
            \path [-] (1) edge (2);
            \path [-] (3) edge (1);
        \end{tikzpicture}
        & 
        \begin{tikzpicture}
            \node[shape=circle,draw=black] (1) at (0,1) {$d_1$};
            \node[shape=circle,draw=black] (2) at (1,0.5) {$d_2$};
            \node[shape=circle,draw=black] (3) at (0,0) {$d_3$};
            \node[shape=circle,draw=black] (4) at (-1,0.5) {$d_4$};
        
            \path [-] (1) edge (4);
            \path [-] (1) edge (2);
            \path [-] (1) edge (3);
            \path [-] (2) edge (3);
        \end{tikzpicture}
        \\
        $[d_1, d_2, d_3, d_4]\rightarrow[d_2,d_3,d_1,d_4]$
        & 
        $[d_3, d_2, d_4, d_1]\rightarrow[d_1,d_2,d_3,d_4]$
    \end{tabular}
    
    \vspace*{1mm}
    \caption{Examples of permutation graphs.}
    \label{fig:permutationgraph}
\end{figure}

%% file: sections/02-related.tex

\section{Background and Related Work}
\label{sec:related} 

\myparagraph{Fairness}
Widely used ranking algorithms at the core of many online systems such as search engines and recommender systems raise fairness considerations, since these algorithms can directly control the exposure each item receives and potentially have societal impacts~\citep{baeza2018bias}.

Similar to other areas of machine learning, various approaches have been proposed to evaluate fairness in ranking.
\citet{zehlike2021fairness} distinguish two lines of work based on how fairness is measured for a ranking policy: probability-based approaches determine the probability that a given ranking is generated by a fair policy~\citep{yang2017measuring, zehlike2017fa, asudeh2019designing, celis2020interventions, celis2017ranking, geyik2019fairness, stoyanovich2018online}, while exposure-based methods aim to ensure that the expected exposure is fairly distributed among items (individual fairness), or item groups (group fairness)~\citep{biega2018equity, mehrotra2018towards,  morik2020controlling, sapiezynski2019quantifying, singh2018fairness, singh2019policy, yadav2019fair, diaz2020evaluating, heuss-2022-fairness, sarvi2021understanding}. 

From the methods belonging to the second category, many have followed the concept of demographic parity, which enforces a proportional allocation of exposure between groups~\citep{yang2017measuring,celis2017ranking,geyik2019fairness,zehlike2017fa}. 
This approach to fairness does not consider merit and only takes into account the group size. 
On the other hand, disparate treatment is a merit-based approach to fair ranking proposed in~\citep{singh2018fairness} which makes exposure allocation dependent on the merit of each group. 
They developed a framework which allows for other forms of fairness definitions that can be formulated as linear constraints. 

Methods introduced in these publications are dependent on their notion of fairness.
However, fairness goals can be application specific. 
In this work, we propose a black-box optimization method for fairness in ranking that is able to optimize a ranking policy w.r.t.\ any arbitrary fairness definition.

\myparagraph{Gradient estimators for permutations}
\label{sec:related:reinforce}
The group of all permutations of size $n$ is called the \emph{symmetric group} and is denoted by $S_n$.
Permutation optimization can be stated as follows:
\begin{equation*}
    \min f(b),\quad b \in S_n,
\end{equation*}
\noindent
where $f:S_n\rightarrow \mathbb{R}$ can be any general function on permutations.
For this combinatorial optimization problem, the gradient solution is not as clear as continuous differentiable optimization.
Instead, a policy-based~\ac{RL} approach can be used to optimize the expectation of $f(b)$ over a differentiable probability distribution $P(b \mid \theta)$, represented by a vector of continuous parameters $\theta$~\citep{berny2000selection}:
\begin{equation}
    \label{eq:expectationoptimization}
    F(\theta) = \mExpect{P(b \mid \theta)}{f(b)}.
\end{equation}
$F(\theta)$ is optimized when $P(b \mid \theta)$ is totally concentrated on $b^*$, the optimum solution of $f(b)$.
For Eq.~\eqref{eq:expectationoptimization}, the REINFORCE estimator~\citep{williams1992simple} can be used:
\begin{equation}
    \label{eq:reinforcegradient}
    \nabla_\theta F(\theta) = \mExpect{P(b \mid \theta)}{f(b) \cdot \nabla_\theta \log P\left(b \mid \theta\right)}.
\end{equation}
\noindent
Finally, since $S_n$ is exponentially large, the expectation in Eq.~\eqref{eq:reinforcegradient} can be estimated through~\ac{MC} sampling:
\begin{equation}
    \label{eq:mcreinforcegradient}
    \nabla_\theta F(\theta) \approx \frac{1}{\lambda} \sum_{i=1}^{\lambda} {f(b_i) \cdot \nabla_\theta \log P\left(b_i \mid \theta\right)},
\end{equation}
\noindent
where $b_1,\dots,b_\lambda \in S_n$ are samples drawn from $P(b \mid \theta)$.

It remains to discuss the probability distribution on $S_n$.
For a thorough exploration of probability distributions for permutations we refer the reader to~\citep{flinger1988multistage}.
The~\ac{PL} model~\citep{luce1959individual, plackett1975analysis} is by far the most widely used permutation distribution in the REINFORCE algorithm, both in general permutation optimization~\citep{gadetsky2020low} and the fairness literature~\citep{singh2019policy,oosterhuis2021computationally}.
\Ac{PL} is represented by a parameter vector $\theta \in \mathbb{R}^n$, and the probability of a permutation $b=[b_1,\dots,b_n]\in S_n$ under PL is calculated as:
\begin{equation}
    \label{eq:plprob}
    P(b \mid \theta) = \prod_{i=1}^{n-1} \frac{\theta_{b_i}}{\sum_{j=i}^{n} \theta_{b_j}}.
\end{equation}
\noindent
Sampling from PL and estimating the log derivative of the probability $P(b \mid \theta)$ is usually done using the Gumbel Softmax trick~\citep{gumbel1954statistical, maddison2014a}.

In this work, we use the REINFORCE algorithm but with our novel~\ac{PPG} distribution instead of \ac{PL}.

\myparagraph{Tabular search}
\label{sec:related:tabular}
In practice, not all the queries are served with a feature-based LTR model.
Tabular models usually achieve optimal performance for head and torso queries for which enough user interactions are available~\citep{NIPS2016_51ef186e}.
In particular, underperforming torso queries are given to the tabular models for a better \emph{memorized} ranking~\citep{zoghi2016click, 10.1145/2911451.2926725, grainger2021ai, Mavridis2020BeyondAR, trotman2017architecture}.
The online LTR literature is filled with methods for tabular model optimization~\citep{pmlr-v48-katariya16,NIPS2016_51ef186e,pmlr-v115-li20b}.
As tabular models are not limited by the capacity of the LTR model, they usually converge to extremely high ranking performance~\citep{zoghi2016click,oosterhuis2021robust}.

This work relates to tabular search by exploiting its results as a use case.
We show in our experiments that, given accurate estimates of utility measures of the items, e.g., their relevance labels, our proposed method for fairness optimization performs significantly better than state-of-the-art fairness optimizers.
Tabular models are practical and important evidence to show that the availability of accurate relevance estimates is not just hypothetical.

%% file: sections/03-method.tex

\section{\ourmethodtitle}
\label{sec:method}
Given a list of items $D$, together with a black-box objective function $f:S_n\rightarrow \mathbb{R}$ that can be queried for every permutation of $D$, our goal is to find a distribution over permutations that optimizes $f$.
We assume that the utility of items, e.g., their relevance level, is given, or an estimate of the utility is obtainable using a~\ac{LTR} model.
Therefore, our task is to post-process a ranked list and output a permutation, or a sequence of permutations, that optimize the given objective function.
To do so, we use the REINFORCE algorithm to search the permutation space and update the PPG distribution parameters.
Below, we first formally define PPG and provide a log derivative formula for the PPG distribution.
After that, we propose a method for efficiently sampling from the PPG distribution.
Finally, we discuss pairwise constraints, including practical examples.

\subsection{\protect{\ac{PPG}} distributions}
\label{sec:method:ppgdistribution}

We define~\ac{PPG} to be a probabilistic graph from which permutation graphs are sampled.

\begin{definition}[\Acf{PPG}]
\label{def:ppg}
    Given a list of items $D$ and a reference permutation $\pi_0$ on $D$, a~\ac{PPG} corresponding to $\pi_0$ is a weighted complete graph $G=(D,E,w)$, whose edges are weighted by a probability value obtained from $w:E\rightarrow[0,1]$.
    For each edge $e\in E$, its weight $w(e)$ indicates the Bernoulli sampling probability that $e$ is included in a permutation graph over $\pi_0$ sampled from $G$.
\end{definition}

\noindent%
In what follows, we write $e_{ij}$ for the edge $(d_i,d_j)$  and $w_{e_{ij}}$ for its weight (or $w_{ij}$ when no confusion is possible).
According to the above definition,~\ac{PPG} represents a permutation distribution.
To sample a permutation from a given PPG, a Bernoulli sampling process is run on the edges of $G$ with their corresponding weights.
Suppose that, after edge sampling, $E_\pi \subset E$ is the set of edges that are selected (i.e., positively sampled) by the sampler and $E\setminus E_\pi$ is the set of remaining, left out edges (i.e., negatively sampled).
The probability of this outcome is calculated as:
\begin{equation}
    P(E_\pi \mid w) = \prod_{e\in E_\pi}w_e \cdot \prod_{e\in E\setminus E_\pi}\bigl(1 - w_e\bigr).
\end{equation}
\noindent
Now, we notice that the set of all permutation graphs over a list of items $D$ is only a small subset of all possible graphs.
This means that $E_\pi$ may not correspond to the edges of a valid permutation graph.
If that happens, i.e., if the graph constructed from the $E_\pi$ edge set is not a permutation graph, we drop $E_\pi$ and repeat the sampling process until we find a valid permutation graph.
Checking if a given graph is a permutation graph or not is possible in linear time~\citep{McConnell1999Modular}.
We write \permspace for the space of all permutation graphs.
Then, the probability that a randomly sampled graph from $G$ is a permutation graph can be computed in theory by:
\begin{equation}
    \label{eq:validpermutationprob}
    \permcond = \sum_{E_\pi\in \permspace} P(E_\pi \mid w).
\end{equation}
Note that in practice, computing $\permcond$ requires an exponential $n!$ number of computations and, hence, is not feasible.
We will come back to this point in Appendix~\ref{sec:logderivativapprox} below.
Using $\permcond$, the permutation probability mass can be written as:
\begin{equation}
    \label{eq:permutationprobabilitymass}
    P(E_\pi \mid w, E_\pi\in \permspace) = \frac{1}{\permcond}\prod_{e\in E_\pi}w_e \cdot \prod_{e\in E\setminus E_\pi}\bigl(1 - w_e\bigr).
\end{equation}
It is not hard to show that repeatedly sampling graphs until we sample a valid permutation graph will lead to the conditional probability distribution of Eq.~\eqref{eq:permutationprobabilitymass}.
We interchangeably use $E_\pi \in \permspace$ for both the positively sampled edges of the permutation graph, and the permutation corresponding to the graph.

To work with the REINFORCE algorithm (Sec.~\ref{sec:related:reinforce}), we need to compute the log derivative of the probability distribution of PPG.
In what follows we assume that $E_\pi \in \permspace$ and drop the conditions from our notation for brevity.
We start by using Eq.~\eqref{eq:permutationprobabilitymass} and taking the log derivative as follows:
\begin{equation}
    \label{eq:ppggradfull}
    \begin{split}
        \frac{\partial \log P(E_\pi)}{\partial w_e} = {}&\frac{\partial}{\partial w_e} \log \left(\prod_{e\in E_\pi}w_e \cdot \prod_{e\in E\setminus E_\pi}\bigl(1 - w_e\bigr)\right) \\
        & - \frac{\partial}{\partial w_e} \log \permcond \\
        = {}&{\frac{\mathbb{I}\bigl(e\in E_\pi\bigr) - w_e}{w_e \cdot \bigl(1 - w_e\bigr)}}
        - {\frac{1}{\permcond}\frac{\partial \permcond}{\partial w_e}}.
    \end{split}
\end{equation}
where $\mathbb{I}(\cdot)$ is the indicator function.
In Appendix~\ref{sec:logderivativapprox} we show the second term can be ignored to obtain the following approximation:
\begin{equation}
    \label{eq:ppggrad}
        \frac{\partial \log P(E_\pi)}{\partial w_e} \approx \frac{\mathbb{I}\bigl(e\in E_\pi\bigr) - w_e}{w_e \cdot \bigl(1 - w_e \bigr)}.
\end{equation}
\noindent
So far, we have defined the PPG and provided the log derivative of its probability distribution.
However, repeatedly dropping the sampled graphs until a valid permutation graph is found, is not efficient.
Next, we present an efficient method for sampling permutations from the PPG distribution.

\subsection{Efficient sampling of permutation graphs}
\label{sec:method:samplingpermutationgraph}
In order to avoid redundantly sampling non-permutation graphs from a PPG, we propose a divide and conquer approach for efficiently constructing a valid permutation graph.
The idea is to divide the list of items into two sub-lists, obtain a permutation on each of the sub-lists by recursively sampling from the given PPG, and merge the two sampled sub-lists into a single permuted list.
Below, we discuss our \emph{merge sampling} method in more detail, but first, let us give an example to illustrate the general idea.

\input{sections/figure_tex/03-fig-samplig}

\begin{example}
    \label{exam:sampling}
    Suppose a reference permutation $[d_1, d_2, d_3, d_4]$ and a PPG are given.
    In order to sample a permutation, we proceed as follows:
    \begin{enumerate}[label=(\Roman*),leftmargin=*]
        \item {\bfseries \emph{Divide.}} We divide the list into two sub-lists $[d_1, d_2]$ and $[d_3,d_4]$.
        \item {\bfseries \emph{Sample sub-lists.}} Each sub-list is sampled independently according to PPG.
        Here, each pair of $[d_1, d_2]$ and $[d_3,d_4]$ are swapped with a probability equal to their weight, i.e., $w_{12}$ and $w_{34}$.
        Suppose only the first pair is swapped $[d_1, d_2]\rightarrow [d_2,d_1]$.
        \item {\bfseries \emph{Merge.}} Finally, we merge $[d_2,d_1]$ and $[d_3,d_4]$.
        In doing so, we keep the order of items in each sub-list unchanged, i.e., no new edges are added between the items of the same sub-list.
        We start from the last item of the first sub-list, $d_1$.
        $d_1$ can go to three possible positions as depicted in Fig.~\ref{fig:samplingexample}: 
        \begin{enumerate*}[label=(Case \arabic*)]
            \item Before $d_3$, so $e_{13}$ is negatively sampled. In this case, since the order of the first sub-list should remain unchanged, $d_2$ has only one possible position.
            \item Between $d_3$ and $d_4$, so only $e_{13}$ is positively sampled. In this case, $d_2$ has two possible positions: before and after $d_3$.
            \item After $d_4$, so both $e_{13}$ and $e_{14}$ are positively sampled and $d_2$ has three possible positions.
            These three possibilities are tested sequentially.
            First, we sample $e_{13}$.
            If it is not selected, it means $d_1$ should remain before $d_3$ and we are done (Case 1).
            Otherwise, $d_1$ should go after $d_3$.
            Then, we sample $e_{14}$ and stop if it is not selected (Case 2).
            Finally, if $e_{14}$ is also selected, we have case 3.
            After $d_1$ has been merged, we repeat the above process for merging $d_2$.
        \end{enumerate*}
        \item {\bfseries \emph{Probability correction.}} If $e_{13}$ is negatively sampled, neither $e_{14}$ or $e_{23}$ can be sampled (Fig.~\ref{fig:samplingexample}, Case 1).
        Therefore, when sampling $e_{13}$, we have to account for the impossible outcome of not selecting $e_{13}$ but selecting at least one of $e_{14}$ or $e_{23}$.
        The probability of this event is calculated as:
    \begin{equation*}
        q_{13}=(1 - w_{13})(w_{14}+w_{23}-w_{14}w_{23}).
    \end{equation*}
    The sum of possible outcomes would then be equal to $1-q_{13}$ and the sampling probability of $e_{13}$ should be corrected by $w_{13}/\bigl(1-q_{13}\bigr)$.
    Similar arguments can be given for Case 2, where $e_{13}$ is positively sampled.
    Similarly, not selecting $e_{14}$ but selecting $e_{24}$ is impossible.
    So, sampling of $e_{14}$ should be done by a probability equal to $w_{14}/\bigl(1-q_{14}\bigr)$, where $q_{14}=(1 - w_{14})w_{24}$.
\hfill $\square$
    \end{enumerate}
\end{example}

\noindent%
In the above example, we saw the simplest case of merging two sub-lists of size two.
More generally, assume we want to merge two permuted sub-lists $[d_1, \dots, d_T]$ and $[d_{T+1}, \dots, d_n]$.
Here we re-indexed the items to simplify the notation.
We start by merging the last item from the top sub-list $d_{T}$ into the bottom sub-list.
Assume $d_{t+1}$ from the top sub-list is merged just after $d_{b}$ in the bottom sub-list.
Now, for merging $d_{t}$, we know that it should be placed before $d_{t+1}$, meaning that at most it can be inverted with $d_{b}$, but not any item after that in the bottom sub-list.
Starting from $d_{T+1}$ on the bottom sub-list, we sample the inversions using the corrected probabilities and stop as soon as one inversion was sampled negatively.
Assume the inversions of $d_t$ with $d_{T+1}, \dots, d_{i-1}$ were positively sampled and we want to sample the inversion with $d_{i}$.
There are two independent impossible outcomes: $e_{ti}$ is sampled negatively, but
\begin{enumerate*}[label=($\Alph*$)]
    \item at least one of the $e_{tj}$ for $i<j\leq b$ is sampled positively; or
    \item at least one of the $e_{t'i}$ for $t'<t$ is sampled positively.
\end{enumerate*}
The probability of these impossibilities can easily be calculated.
We write $q_{ti}$ for the probability of an impossible outcome due to negatively sampling $e_{ti}$:\looseness=-1
\begin{equation}
    \label{eq:impossibleprobability}
    q_{ti} = \bigl(1 - w_{ti}\bigr)\bigl(P(A)+P(B)-P(A)P(B)\bigr).
\end{equation}
As the sum of all possible events is $1-q_{ti}$, the inversion of $e_{ti}$ should be sampled with the corrected probability of $w_{ti}/\bigl(1-q_{ti}\bigr)$.

\input{sections/figure_tex/03-alg-sampling}

The recursive algorithm of sampling a permutation graph is shown in Algorithm~\ref{alg:sampling}.
The main part of the algorithm is the {\bfseries Merge} function in line 14.
The pseudo-code for the merge function is given in Algorithm~\ref{alg:merging}.
The worst-case complexity of calculating $q_{ti}$ by Eq.~\eqref{eq:impossibleprobability} is $\mathcal{O}(n)$ which makes the worst-case complexity of {\bfseries Merge} (Algorithm~\ref{alg:merging}) equal to $\mathcal{O}(n^3)$.
In practice, we update the reference permutation each time a better permutation with respect to the objective function was sampled during training.
As the training goes on, the reference permutation gets closer to the optimum permutation and the model gets more confident in it.
This means that the inversion probabilities are constantly decreasing.
In the extreme case when the probability of changing the reference permutation becomes very small, the average complexity of {\bfseries Merge} becomes linear and the total complexity of {\bfseries Sample} (Algorithm~\ref{alg:sampling}) becomes equal to $\mathcal{O}(n \log{} n)$.

Experiments show that our efficient sampling method does not provide a uniform sampling: in the trade-off between accurately sampling according to a given PPG distribution and having an efficient sampler, our method is inclined to the latter.
Our experimental results in Sec.~\ref{sec:results:performance} verify that this sampling method is effective in fairness optimization.
Further analyzing the accuracy-efficiency trade-off of sampling PPG distributions and proposing sampling methods with different degrees of accuracy and efficiency is an interesting direction that we leave for future work.

\subsection{Learning PPG weights}
We use the REINFORCE algorithm to train the weights of our PPG distribution as described in Sec.~\ref{sec:related:reinforce}.
For~\ac{MC} sampling from the permutation distribution as required in Eq.~\eqref{eq:mcreinforcegradient}, we use the {\bfseries Sample} algorithm as described in Algorithm~\ref{alg:sampling}.
The only difference with the standard REINFORCE is that we keep track of the minimum value for the objective function and the corresponding permutation.
After each permutation has been sampled and the objective function has been calculated, we update the reference permutation with the best sampled permutation.
\input{sections/figure_tex/03-alg-learning}
Algorithm~\ref{alg:learning} contains the pseudo-code for learning the PPG weights.
In line 8, $P$ is the permutation matrix of $E_{\pi_i}$.

\subsection{Pairwise constraints}
\label{sec:method:constraints}
A good property of~\ourmethod is that pairwise constraints on the permutations can be handled without extra computational complexity.
Pairwise constraints in the form of forbidden inversions, can be handled by setting to zero the appropriate set of edges in the~\ac{PPG}.
Here we give some practical examples of this type.

\myparagraph{Intra-group fixed-ranking}
An intra-group constraint ensures that the ranking of a group of items remains unchanged among themselves.
Consider, for example, $D=[d_1, d_2, d_3, d_4]$ with the group $g=\{d_2, d_3, d_4\}$.
An intra-group fixed-ranking constraint on $g$ means that the ranking of the items within $g$ should remain the same as the reference permutation.
The permutation $[d_2,d_1,d_3,d_4]$ meets this constraint, but in $[d_1,d_3,d_2,d_4]$ or $[d_4,d_1,d_3,d_2]$ the ranking of items of $g$ is changed and the constraint is violated.
Multiple groups can be defined over the item list, and the groups need not be disjoint.
To handle intra-group constraints, in the initial~\ac{PPG}, we simply set to zero the weights of all the edges between items within the same group.
This ensures that none of these edges will be sampled during the learning process.
As a result, no inversion is performed over the items within the same group.

A use case for this constraint is in group fairness, where we post-process the output of an \ac{LTR} algorithm to make the ranking fair with respect to some grouping.
Each inversion in the output of the LTR algorithm means degradation in our estimated best ranking.
Therefore, the inversions are performed only to improve fairness.
But inverting two items from the same fairness group does not change the fairness metric.
Such inversions that degrade the ranking and do not improve fairness can be avoided by setting an intra-group constraint on the permutations.

\myparagraph{Inter-group fixed-ranking}
Inter-group constraints can be used to ensure that one group is ranked higher than the other group.
Consider, for example, $D=[d_1, d_2, d_3, d_4, d_5]$ with the groups $g_1=\{d_1, d_2\}$ and $g_2=\{d_4, d_5\}$.
The permutation $[d_2, d_3, d_1, d_4, d_5]$ meets the inter-group constraint, as all the items of $g_1$ are ranked higher than all the items of $g_2$, the same as the reference permutation.
But $[d_1, d_4, d_3, d_2, d_5]$ violates the constraint because $d_4 \in g_2$ is ranked higher than $d_2 \in g_1$.
To handle this constraint, in the initial~\ac{PPG}, we set to zero the weights of all edges whose endpoints are in different groups.
Consequently, there would be no inversion between two items from different groups.

A use case for this constraint is in amortized fairness, where fairness is measured over multiple sessions (with the same or different queries).
Since the objective function is calculated over multiple lists, in~\ourmethod we can concatenate the lists and search for a solution over the concatenated list.
In this case, items from different sessions should not be inverted: we are only allowed to change the permutation within each session.
Using an inter-group fixed-ranking constraint, with each session considered as a group, is a simple solution for such a use case.

\myparagraph{Time-aware ranking}
In some scenarios, such as news search or job search, items are associated with time and it is sometimes important not to rank very old items before recent items.
A time-aware ranking constraint is a special case of inter-group fixed-ranking, where groups are defined based on time.

\myparagraph{Context-aware ranking}
Another special case of an inter-group fixed-ranking constraint is context-aware ranking.
Assume, for example, the case of sponsored links.
We want to post-process the ranking to make it fair, but in addition to that, we want to have the relevant sponsored links to appear at the top of the list.
We can define groups based on the sponsored status of items and set an inter-group constraint on the PPG graph.

We use the first two examples, namely intra-group and inter-group fixed-ranking, in our experiments.
With the intra-group constraint, we control the ranking performance of permutations while minimizing the fairness, and with the inter-group constraint, we train over multiple sessions of one query with a single PPG model.

%% file: sections/figure_tex/03-fig-samplig.tex

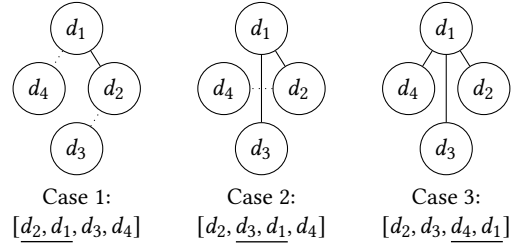
\begin{figure}
    \begin{tabular}{ccccc}
        \begin{tikzpicture}
            \node[shape=circle,draw=black] (1) at (0,1.6) {$d_1$};
            \node[shape=circle,draw=black] (2) at (0.5,0.8) {$d_2$};
            \node[shape=circle,draw=black] (3) at (0,0) {$d_3$};
            \node[shape=circle,draw=black] (4) at (-0.5,0.8) {$d_4$};
        
            \path [-] (1) edge (2);
            \path [dotted] (1) edge (4);
            \path [dotted] (3) edge (2);
        \end{tikzpicture}
        & &
        \begin{tikzpicture}
            \node[shape=circle,draw=black] (1) at (0,1.6) {$d_1$};
            \node[shape=circle,draw=black] (2) at (0.5,0.8) {$d_2$};
            \node[shape=circle,draw=black] (3) at (0,0) {$d_3$};
            \node[shape=circle,draw=black] (4) at (-0.5,0.8) {$d_4$};
        
            \path [-] (1) edge (2);
            \path [-] (3) edge (1);
            \path [dotted] (2) edge (4);
        \end{tikzpicture}
        & &
        \begin{tikzpicture}
            \node[shape=circle,draw=black] (1) at (0,1.6) {$d_1$};
            \node[shape=circle,draw=black] (2) at (0.5,0.8) {$d_2$};
            \node[shape=circle,draw=black] (3) at (0,0) {$d_3$};
            \node[shape=circle,draw=black] (4) at (-0.5,0.8) {$d_4$};
        
            \path [-] (1) edge (4);
            \path [-] (1) edge (3);
            \path [-] (1) edge (2);
        \end{tikzpicture}
        \\
        Case 1: 
        &  &
        Case 2: 
        &  &
        Case 3: 
        \\
        $[\underline{d_2,d_1},d_3,d_4]$
        &  &
        $[d_2,\underline{d_3,d_1},d_4]$
        &  &
        $[d_2,d_3,\underline{d_4,d_1}]$
    \end{tabular}
    
    \vspace*{1mm}
    \caption{Different cases in Example~\ref{exam:sampling}. Dotted lines show the impossible-to-be-selected edges due to the position of $d_1$ in the bottom sub-list.}
    \label{fig:samplingexample}
\end{figure}

%% file: sections/figure_tex/03-alg-sampling.tex
\begin{algorithm}[t]
    \caption{{\bfseries ~Sample$(D,G)$.} Sampling Permutation Graph}
    \label{alg:sampling}
    \SetAlgoLined
    \KwIn{$D=[d_1,d_2,\dots,d_n]$, $G=(D,E,w)$}
    \KwOut{$D'$ (a permutation on $D$, sampled from $G$)}
    \If{$n==2$}{
        Bernoulli sample with probability $w_{12}$\\
        \eIf{Positive}{
            Return $[d_2,d_1]$
        }{
            Return $[d_1,d_2]$
        }
    }
    Split $D$ in half to $D_t$ and $D_b$\\
    Set $G_t$ to the upper left square of $G$ corresponding to $D_t$\\
    Set $G_b$ to the lower right square of $G$ corresponding to $D_b$\\
    $D'_t=${\bfseries ~Sample}$(D_t,G_t)$\\
    $D'_b=${\bfseries ~Sample}$(D_b,G_b)$\\
    Return {\bfseries ~Merge}$(D'_t, D'_b, G)$
\end{algorithm}

\begin{algorithm}[t]
    \caption{{\bfseries ~Merge$(D_t, D_b, G)$.} Merging Two Sampled Permutations}
    \label{alg:merging}
    \SetAlgoLined
    \KwIn{$D_t=[d_1, \dots, d_T]$, $D_b=[d_{T+1}, \dots, d_n]$, $G=(D_t\cup D_b,E,w)$}
    \KwOut{$D'$ (a merged permutation on $D_t\cup D_b$, sampled from $G$)}
    Initialize the last merged index $b_{last} = n+1$\\
    \For(){$t$ from $T$ down to $1$}{
        Insert $d_t$ on top of $D_b$\\
        \For(){$i$ from $T+1$ to $b_{last}$}{
            Calculate $q_{ti}$ using Eq.~\eqref{eq:impossibleprobability}\\
            Bernoulli sample with probability $w_{ti}/\bigl(1-q_{ti}\bigr)$\\
            \eIf{Negative}{
                $b_{last} = i$\\
                Break
            }{
                Invert: $[\dots,d_t,d_i,\dots] \rightarrow [\dots,\underline{d_i,d_t}, \dots]$
            }
        }
    }
    Return $D_b$
\end{algorithm}

%% file: sections/figure_tex/03-alg-learning.tex
\begin{algorithm}[t]
    \caption{Learning~\ac{PPG} weights}
    \label{alg:learning}
    \SetAlgoLined
    \KwIn{$\pi_0$, $G=(D,E,w)$, $f$, $\eta$}
    \KwOut{Updated $\pi_0$ and $G$}
    \While(){Not converged}{
        \For{$i=1 \dots \lambda$}{
            Use Algorithm~\ref{alg:sampling} to sample a permutation $E_{\pi_i}$.\\
            Use Eq.~\eqref{eq:ppggrad} to calculate the log derivative of PPG.\\
            \If(){$f(E_{\pi_i}) < f(\pi_0)$}{
                $\pi_0 = E_{\pi_i}$ \\
                Update $G$ accordingly:\\
                $\qquad G=P^T\cdot G\cdot P$
            }
        }
        Use Eq.~\eqref{eq:mcreinforcegradient} to estimate gradients.\\
        Update the weights by gradient descent, using learning rate $\eta$.\\
        
    }
    
    Return $\pi_0$, $G$
\end{algorithm}

%% file: sections/04-experiments.tex

\section{Experimental Setup}
\label{sec:experiments} 
In order to show the effectiveness of~\ourmethod, we perform experiments on real-world public data and compare the performance of different fairness methods.
Below, we detail the datasets, experimental setup, and the baselines that we compare to.

\subsection{Data}

\myparagraph{MSLR}
This is a regular choice in counterfactual \ac{LTR} research~\citep{vardasbi2020when,vardasbi2021mixture,joachims2017unbiased}, as well as fairness studies~\citep{diaz2020evaluating, yadav2021policy, oosterhuis2021computationally}.
We use Fold 1 of MSLR-WEB30k~\citep{qin2013introducing} with 5-level relevance labels.
On average, MSLR has 120.19 documents per query.
We follow~\citep{yadav2021policy} and divide the items into two groups based on their QualityScore2 (feature id 133) with a threshold of $10$.
This gives a 3:2 ratio between the groups' population.
We choose a subset of the MSLR test set for post-processing based on the following criteria.
We filter the queries for which there is not at least one fully relevant item, i.e., level $4$.
We also filter the queries that only contain relevant items from one group, as the DTR metric cannot be used for such queries.
For the remaining queries, we subsample queries with long item lists to have a maximum of $20$ items, based on their LTR score.
The intuition is that usually, a top-$k$ cut of the items are shown to users in online search engines.
This is in line with previous fairness and online LTR studies~\citep[e.g.][]{yadav2021policy, vardasbi2020when, vardasbi2021mixture}.

\myparagraph{TREC}
Our second dataset is the academic search dataset provided by the TREC Fair Ranking track\footnote{https://fair-trec.github.io/} 2019 and 2020~\citep{trec-fair-ranking-2019}.
TREC 2019 and 2020 editions of the dataset come with 632 and 200 train and 635 and 200 test queries, respectively, with an average of 6.7 and 23.5 documents per test query.
This dataset has been previously used in fair ranking research~\citep{ sarvi2021understanding, heuss-2022-fairness, kirnap2021estimation}.
Following~\citep{sarvi2021understanding}, we divide the items (i.e., papers) into two groups based on their authors' h-index.


\subsection{Setup}

\myparagraph{LTR model}
We use a neural network with two hidden layers of width 256, ReLU activations, dropout of $0.1$, and a learning rate of $0.01$.
As it is important to have a calibrated LTR for fairness, as noted in~\citep{diaz2020evaluating}, we use a pointwise loss function in the form of~\ac{MSE}.
We notice that the dynamic range of our LTR model is limited: around $95\%$ of the items are concentrated in an interval of length $0.05$ in all of the datasets.
As both DTR and EEL metrics rely on relevance estimates, to calibrate the scores onto a realistic interval, we min-max normalize the scores per query onto $[0,5)$ and then discretize them to integer grades from $0$ to $4$, similar to the relevance grades in popular LTR datasets~\citep{qin2013introducing}.

\myparagraph{Metrics}
We evaluate models for fair ranking in terms of user utility and item fairness. 
For utility, we use \ac{NDCG} and report NDCG@10.
For fairness we evaluate models using two metrics: DTR~\citep{singh2018fairness} and EEL~\citep{diaz2020evaluating}.

Given two item groups, $G_1$ and $G_2$, DTR, measures how unequal exposure ($exp$) is allocated to the two groups based on their merit, which, in our case, translates to the average utility ($u$) of each group:
\begin{equation}
DTR(G_1, G_2\mid P) = \frac{exp(G_1\mid P)/u(G_1\mid q)}{exp(G_2\mid P)/u(G_2\mid q)}
\end{equation}
where $P$ is the ranking policy.
The optimal DTR value is $1$: each group is exposed proportional to their utility.
Here, as we are minimizing the fairness metric, we always keep the DTR $>1$.
This means that $G_1$ and $G_2$ may change for different queries.

EEL is defined to be the Euclidean distance between the expected exposure each group receives and their target exposure in an ideal scenario where items with the same utility have the same probability of being ranked higher than each other, while all the items with higher utility should always be placed above lower utility items~\citep{diaz2020evaluating}.
The optimal value of EEL is $0$.

\myparagraph{Expectation over sessions}
The fairness metrics DTR and EEL are generally calculated as an expectation over multiple sessions of a query.
For~\ourmethod, when taking the expectation over $N$ sessions, we concatenate the list of items for each query to itself, $N$ times, and set the inter-group fixed-ranking constraint (Sec.~\ref{sec:method:constraints}) to prevent items from different sessions from being mixed up.

\subsection{Baselines}
\ac{PL} optimization is the state-of-the-art method for black-box optimization of fairness metrics~\citep{singh2019policy, oosterhuis2021computationally}, as well as general permutation functions~\citep{gadetsky2020low}.
As PPG is a substitute for PL, the most important baseline in our experiments is PL optimization.\footnote{We use the public implementation of~\cite{gadetsky2020low} with some minor modifications.}
Specific to the EEL metric, an in-processing method is proposed in the original EEL paper~\citep{diaz2020evaluating}, which cannot be compared to our post-processing method.
They also use result randomizations based on PL and show it achieves good fairness-ranking trade-offs.
Therefore, to the best of our knowledge, the PL optimization of EEL is the sole state-of-the-art algorithm available in the literature.

For DTR optimization we compare PPG to the state-of-the-art convex optimization method introduced by~\citet{singh2018fairness}, denoted by FOE. 
This method is a post-processing approach to fair ranking that finds a utility-maximizing marginal probability matrix $P$ that avoids disparate treatment by solving a linear optimiziation problem. To compute the stochastic ranking policy,  it uses the Birkhoff-von Neumann algorithm~\citep{birkhoff-1940-lattice} to decompose $P$ into permutation matrices that correspond to rankings. Following~\citep{sarvi2021understanding} we use two variants of FOE based on hard vs. soft doubly stochastic matrix constraints, and call them $FOE^H$ and $FOE^S$, respectively.\footnote{We used the implementation from \url{https://github.com/MilkaLichtblau/BA_Laura}.}
Similar to PPG, this model generates a stochastic ranking policy that directly optimizes for DTR, making it a suitable baseline to compare with.

In both the DTR and EEL case, we also include the ranking and fairness performance of the LTR model without any post-processing, as well as the performance of randomizing the items, i.e., picking permutations uniformly at random from $S_n$, denoted by RAND.
Given enough sessions per query, randomized rankings usually have outstanding fairness but relatively bad ranking performance.

%% file: sections/05-results.tex

\input{sections/figure_tex/05-fig-performance}

\section{Results}
\label{sec:results} 
We run experiments to address the following research questions:
\begin{enumerate}[label=(RQ\arabic*),leftmargin=*]
    \item Compared to existing methods for ranking fairness optimization, what is the effectiveness of~\ourmethod in finding the fairest ranking for different fairness measures?
    \item How does~\ourmethod perform compared to~\ac{PL} for queries with a small number of repeating sessions?
    \item Can the possibility of pairwise constraints in~\ourmethod be used to control other measures, e.g., ranking performance, while optimizing fairness?
    \item How does an accurate estimate of utility, e.g., the relevance labels, affect different fairness optimization methods?
\end{enumerate}

\subsection{Fairness optimization performance}
\label{sec:results:performance}
We first address (RQ1) and (RQ2) concerning the performance comparison of~\ourmethod to existing fairness optimization methods.
Fig.~\ref{fig:performance} shows a performance comparison of our~\ourmethod and PL for optimizing DTR and EEL on three datasets TREC 2019, TREC 2020, and MSLR, for different numbers of sessions per query.\footnote{We discuss the ``PPG$~+~$intra'' legend in Sec.~\ref{sec:results:pairwise}.}
For both fairness measures, lower means fairer, i.e., better.
We omit FOE$^H$ in this figure for better visualization, as it is surpassed by FOE$^S$ in all three tested datasets.

\myparagraph{EEL}
Regarding (RQ1), we observe in Fig.~\ref{fig:performance} that for EEL, our PPG method performs better than PL on all three datasets, both being far from the fairness obtained from randomized items.
The PL method, after enough sessions, converges to the fairness performance of the LTR output.
In terms of ranking performance when optimizing for EEL, PPG is slightly worse than PL on the MSLR and TREC 2019 datasets.

\myparagraph{DTR}
When optimizing for DTR, PPG and PL both hurt the initial fairness of the LTR output.
Here, FOE$^S$, designed specifically for DTR optimization, has the best performance on all three datasets.
On the MSLR and TREC 2020 datasets, PPG and PL perform closely to the randomized items.
Comparing PPG and PL to each other, PPG is slightly fairer on the TREC 2019 and MSLR datasets.
In the case of DTR optimization for the MSLR dataset, it is interesting to note that randomization slightly hurts fairness of the LTR output, by around $3\%$ relative difference.
FOE$^S$ correctly sticks to the LTR output in this case.
The message from comparing PPG and PL on the three tested datasets is that, PPG leads to slightly fairer performance than PL when optimized for EEL, while the two perform nearly the same when optimized for DTR.

\myparagraph{Small number of sessions}
When the number of sessions is limited (RQ2), we see a decisive advantage for PPG compared to PL in EEL.
For EEL, on all three datasets, PPG converges at the first session, i.e., it finds a deterministic permutation for one session of each query which has a better fairness value than the expected fairness of PL-generated permutations after $32$ sessions.
In this sense, PL can be thought of as the {\bfseries mean$(\cdot)$} aggregator, whereas PPG is the {\bfseries min$(\cdot)$} aggregator; {\bfseries mean$(\cdot)$} needs more sessions to converge to the optimum value.
For DTR, PPG needs $4$ sessions to converge, while PL converges after $8$ sessions.
Therefore, the answer to (RQ2) on our tested datasets is that before a query is repeated many times, PPG is preferred to PL.
The reason is that in PPG the reference permutation is set to the best sampled permutation, so in deterministic scenarios, i.e., with only one session per query, PPG is the go-to method.

\input{sections/figure_tex/05-fig-performance-true}
\subsection{Pairwise constraints}
\label{sec:results:pairwise}
To answer (RQ3) about the effectiveness of pairwise constraints, we set the intra-group fixed-ranking constraint on a PPG model as follows:
The items in each fairness group are constrained to have the same ordering as the LTR output.
This means that the weight of each edge whose endpoints belong to the same fairness group is initialized by zero and consequently will remain zero during training.
This is a conservative way of not hurting the ranking performance too much, while searching for a fair ranking (see Sec.~\ref{sec:method:constraints}).
The ``PPG$~+~$intra'' legend in Fig.~\ref{fig:performance} shows the results.
In this figure we see that in all the cases, the ranking performance, measured by nDCG@10, is noticeably improved by adding the intra-group constraint over PPG.
This ranking performance improvement comes with a negligible cost of fairness degradation:
in the EEL case, ``PPG'' and ``PPG + intra'' have the same fairness performance, while for DTR, there is a slight degradation of less than $3\%$ relative difference in fairness performance on all three datasets.
So we can answer (RQ3) positively: using proper pairwise constraints \emph{does} help to improve other measures, while optimizing for fairness.
This possibility is very useful in scenarios where the true relevance labels are unknown and the exact ranking performance cannot be measured.
Imposing intra-group fixed-ranking constraints is a conservative way of not hurting our best estimation of the ideal ranking.

\subsection{Accurate utility estimates}
One advantage of post-processing for fairness optimization is the possibility of improving future fairness for the torso queries.
In this section, we investigate the effectiveness of different fairness optimization methods for specialized queries where highly confident, accurate utility estimates are available through tabular models.
\looseness=-1

Fig.~\ref{fig:performanceknown} shows the comparison of different fairness optimization methods when accurate relevance labels are known.
Here, we only include the results of ``PPG + intra'', as we have shown its superiority to the simple ``PPG'' in Sec.~\ref{sec:results:pairwise}.
For brevity, we only mention ``PPG'' instead of ``PPG + intra'' in the following.
The gap between the performance of PPG to other baselines is huge, for both fairness and ranking.
PPG is the clear winner in all the tested datasets and both fairness metrics.
For EEL, PPG is able to achieve the optimal value of $0$.
For DTR, PPG comes very close to the optimal value of $1$ on the TREC 2019 and MSLR datasets.
Compared with FOE$^H$ and FOE$^S$, we observe that knowledge of accurate utility estimates helps PL and PPG more and causes them to perform better.
This is in contrast with our observation with noisy LTR outputs in Sec.~\ref{sec:results:performance}.\looseness=-1

In terms of the ranking performance, we observe that the intra-group fixed-ranking has helped PPG to maintain the ranking near ideal, while minimizing the fairness measure.

%% file: sections/figure_tex/05-fig-performance.tex
{
    \setlength{\tabcolsep}{1pt}
    \newcommand{\performancescale}{0.28}
    \newcommand{\fairnesshspace}{2.5em}
    \newcommand{\ndcghspace}{1em}

    \begin{figure*}[t]
        \centering
        \begin{tabular}[]{rcrccrcrc}
    
            \rotatebox[origin=lt]{90}{\hspace{\fairnesshspace} \small DTR}
            &
            \includegraphics[scale=\performancescale]{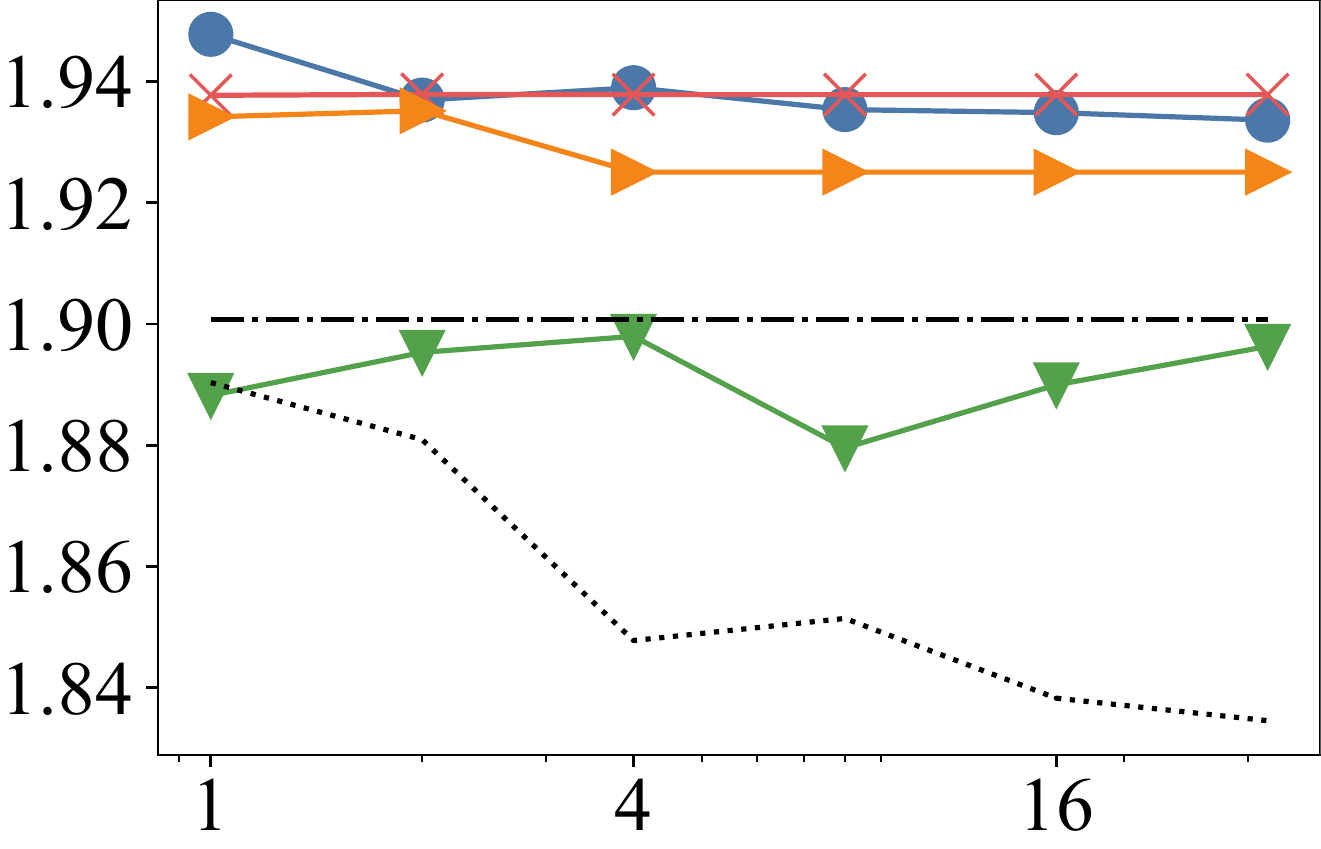}
            &
            \rotatebox[origin=lt]{90}{\hspace{\ndcghspace} \small nDCG@10}
            &
            \includegraphics[scale=\performancescale]{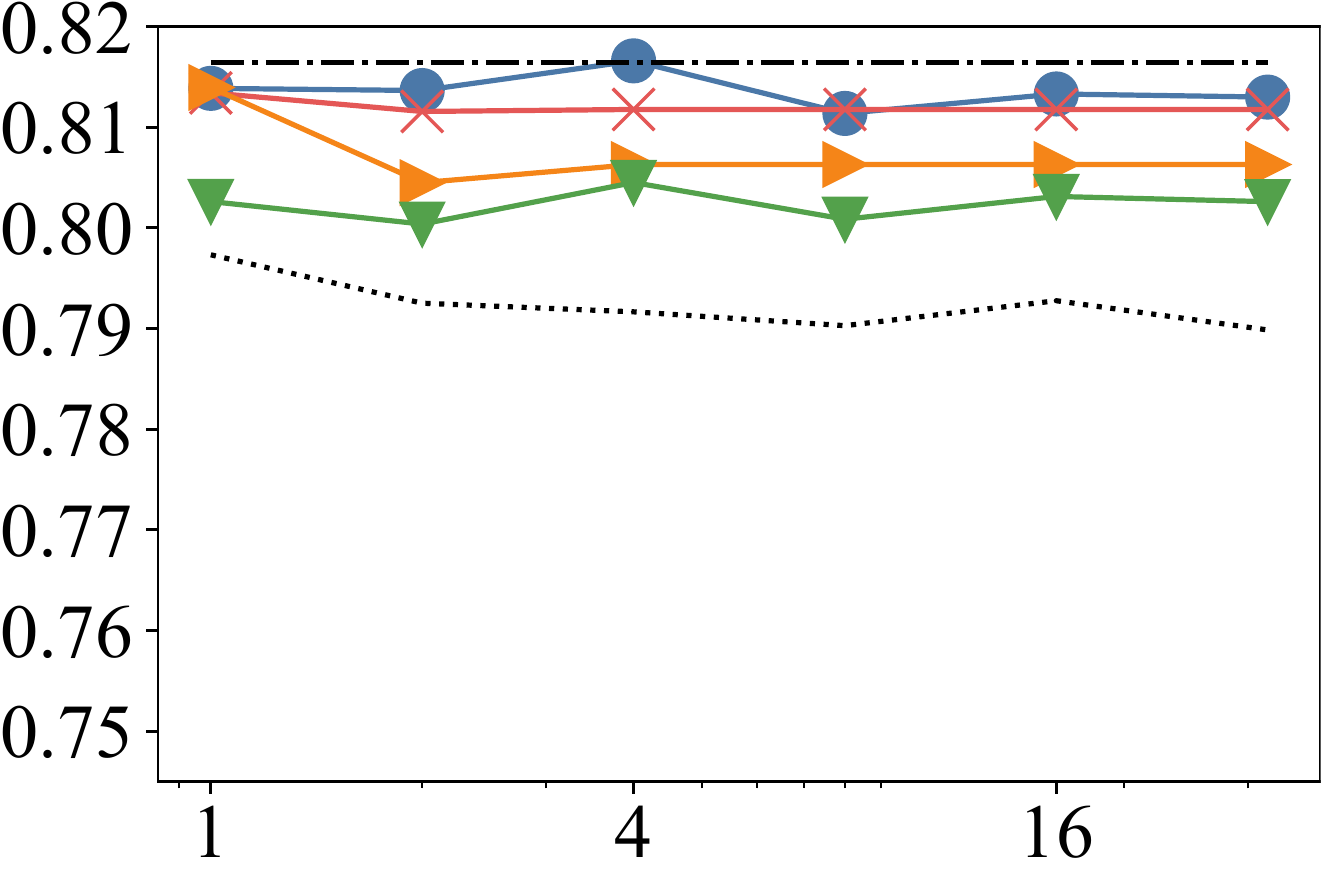} 
            & &
            \rotatebox[origin=lt]{90}{\hspace{\fairnesshspace} \small EEL}
            &
            \includegraphics[scale=\performancescale]{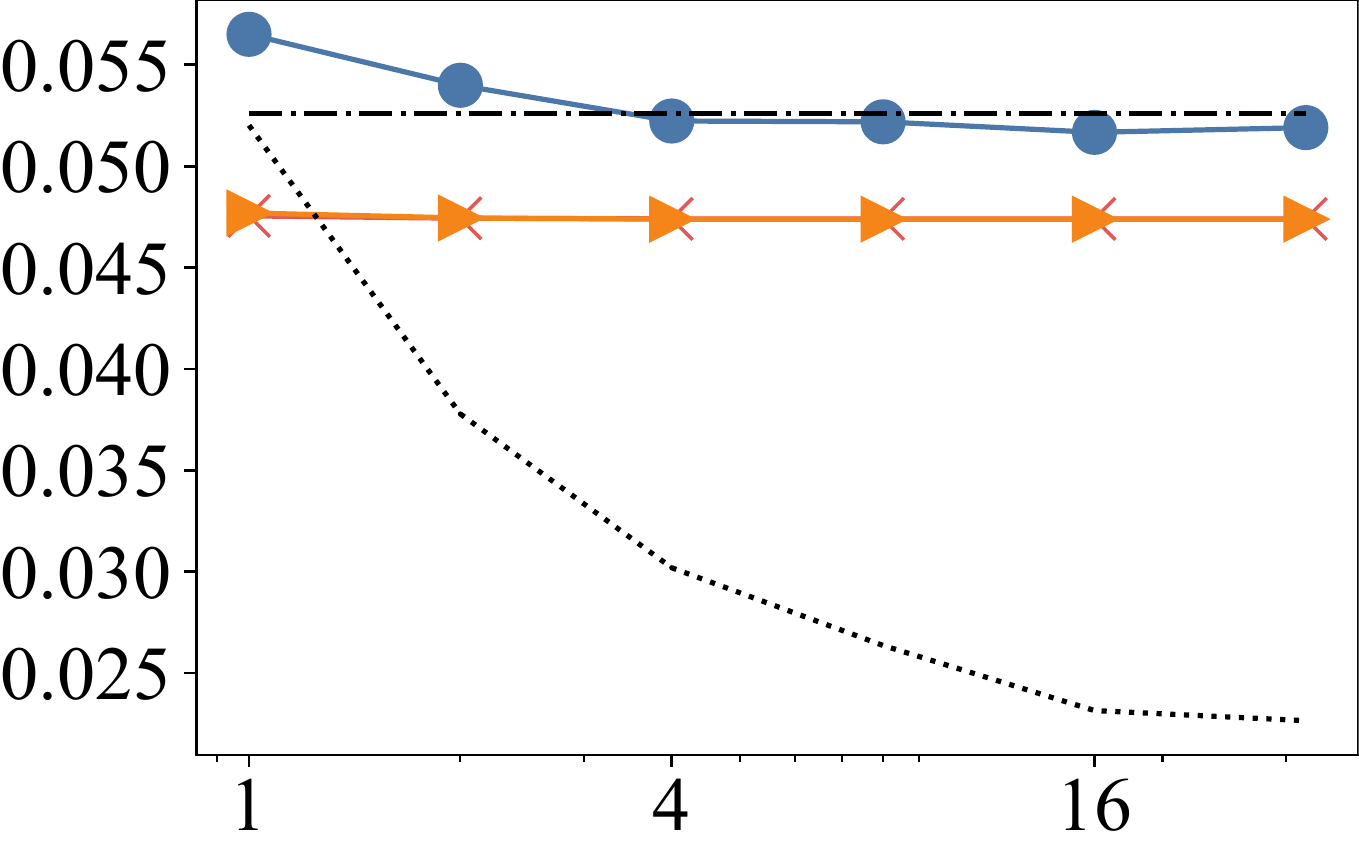}
            &
            \rotatebox[origin=lt]{90}{\hspace{\ndcghspace} \small nDCG@10}
            &
            \includegraphics[scale=\performancescale]{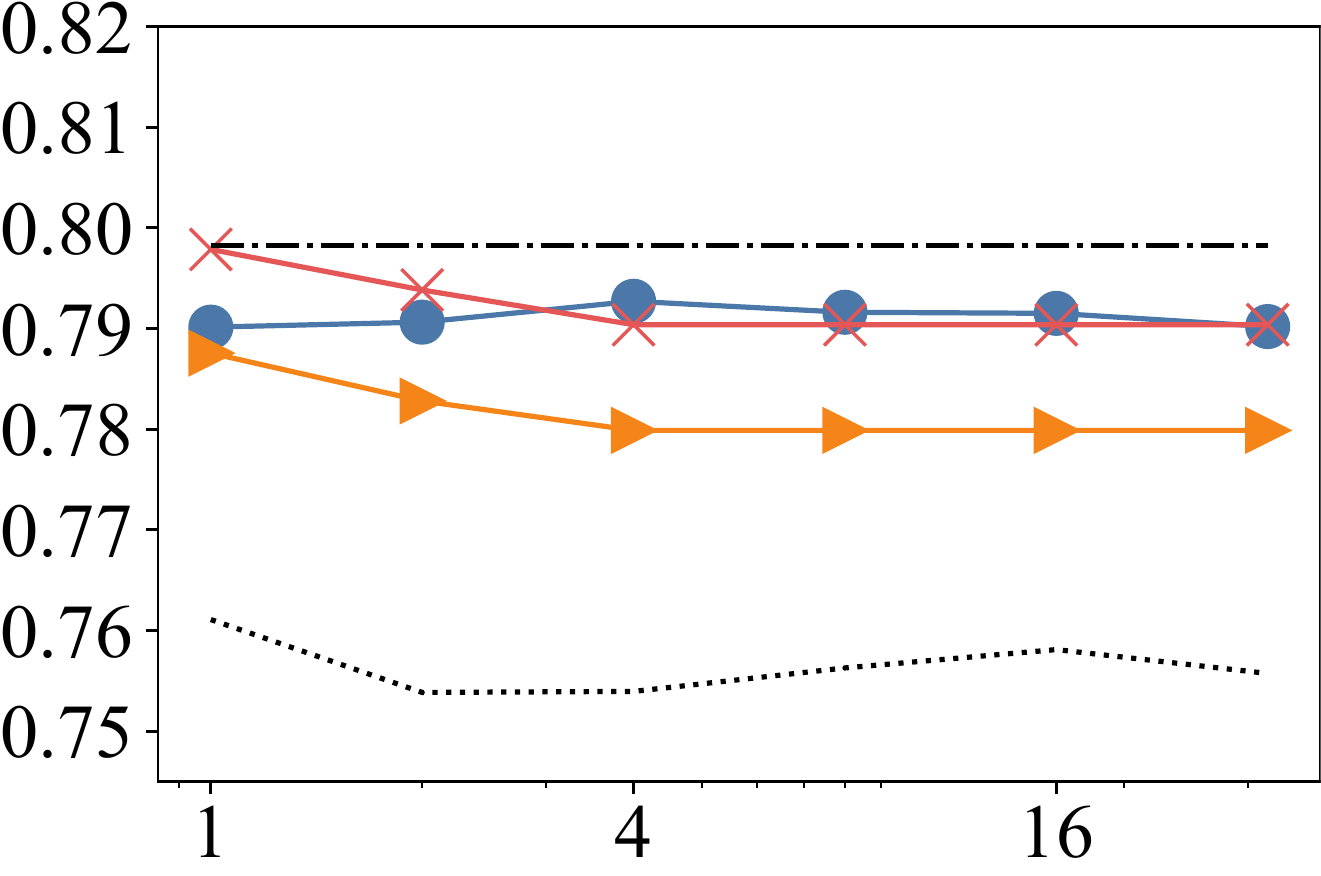} \\
    
            \rotatebox[origin=lt]{90}{\hspace{\fairnesshspace} \small DTR}
            &
            \includegraphics[scale=\performancescale]{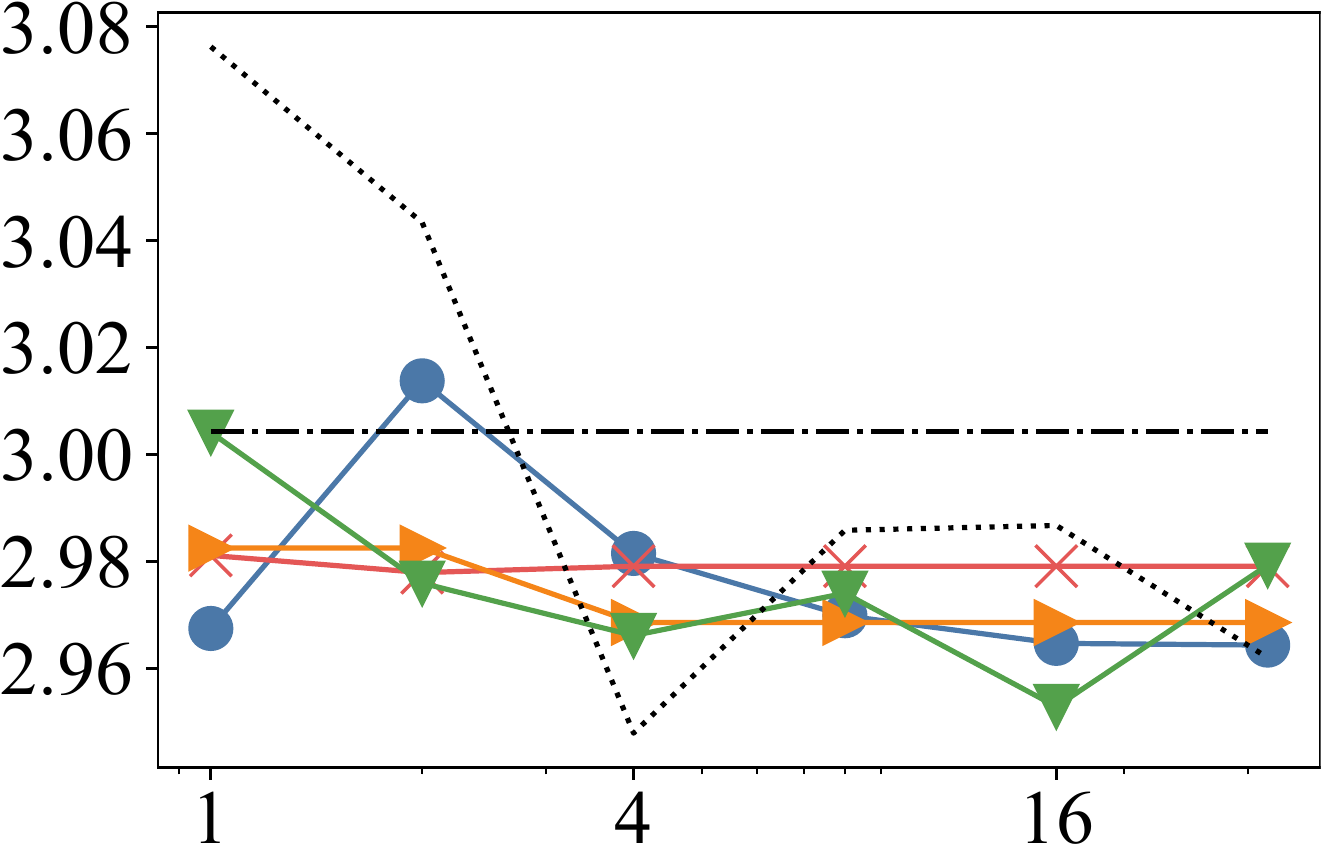}
            &
            \rotatebox[origin=lt]{90}{\hspace{\ndcghspace} \small nDCG@10}
            &
            \includegraphics[scale=\performancescale]{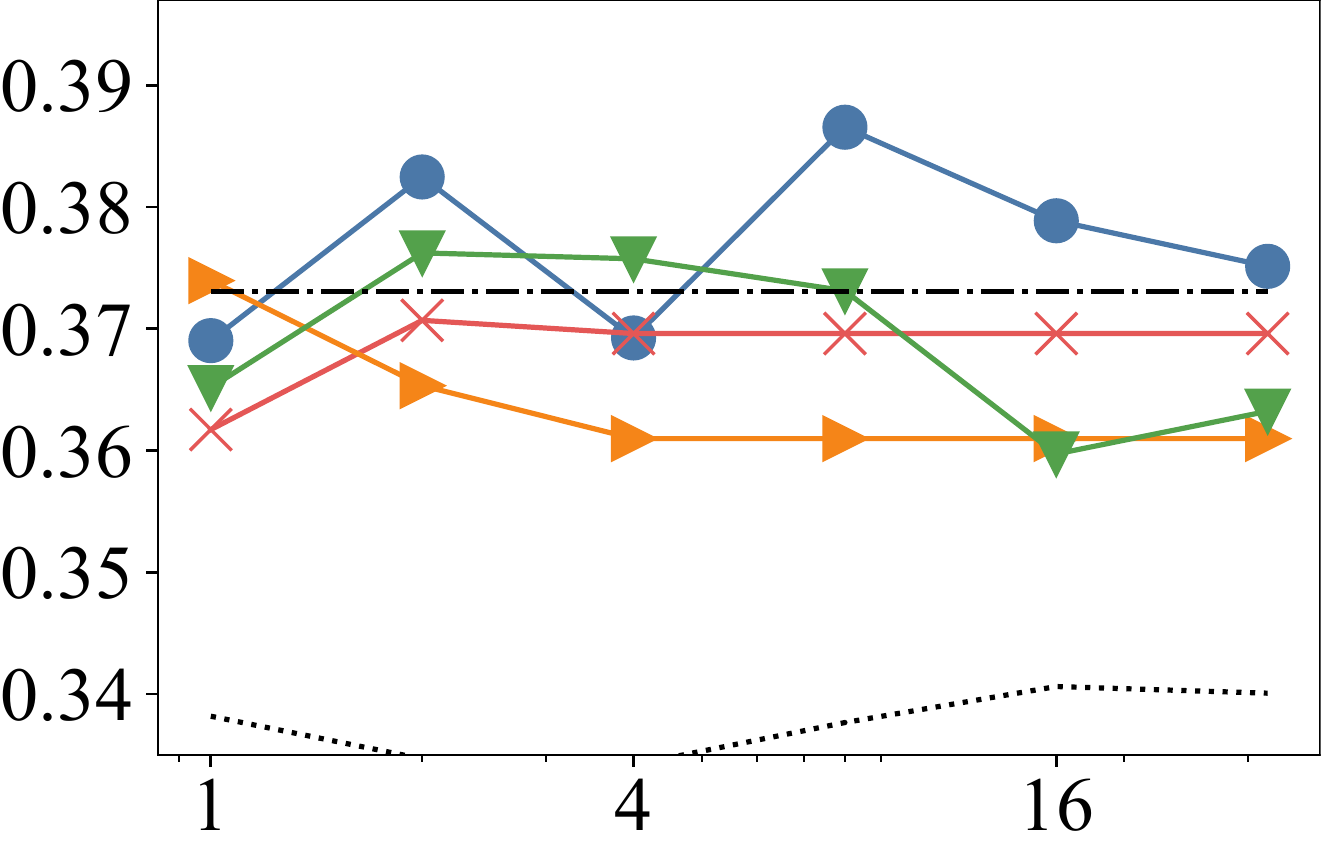} 
            & &
            \rotatebox[origin=lt]{90}{\hspace{\fairnesshspace} \small EEL}
            &
            \includegraphics[scale=\performancescale]{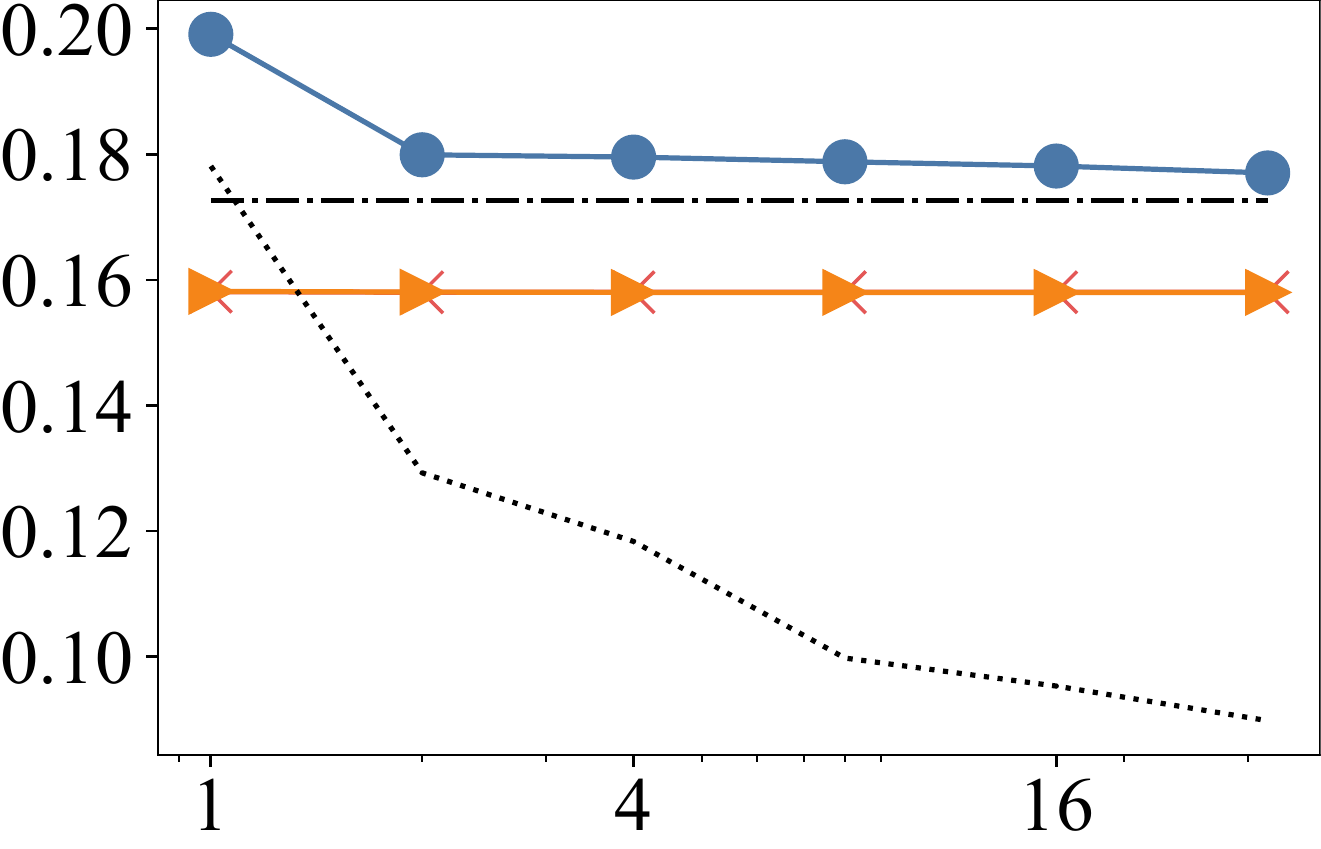}
            &
            \rotatebox[origin=lt]{90}{\hspace{\ndcghspace} \small nDCG@10}
            &
            \includegraphics[scale=\performancescale]{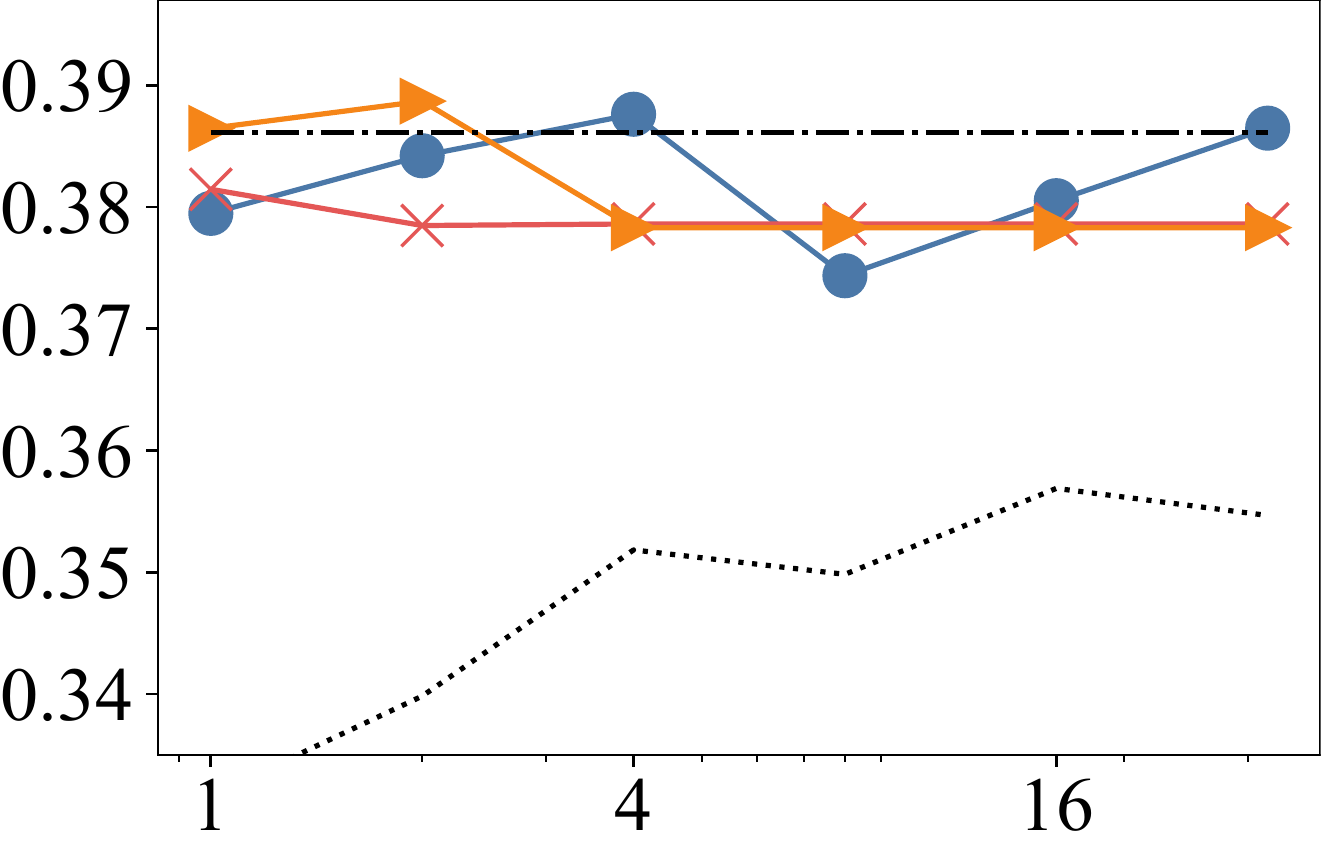} \\
    
            \rotatebox[origin=lt]{90}{\hspace{\fairnesshspace} \small DTR}
            &
            \includegraphics[scale=\performancescale]{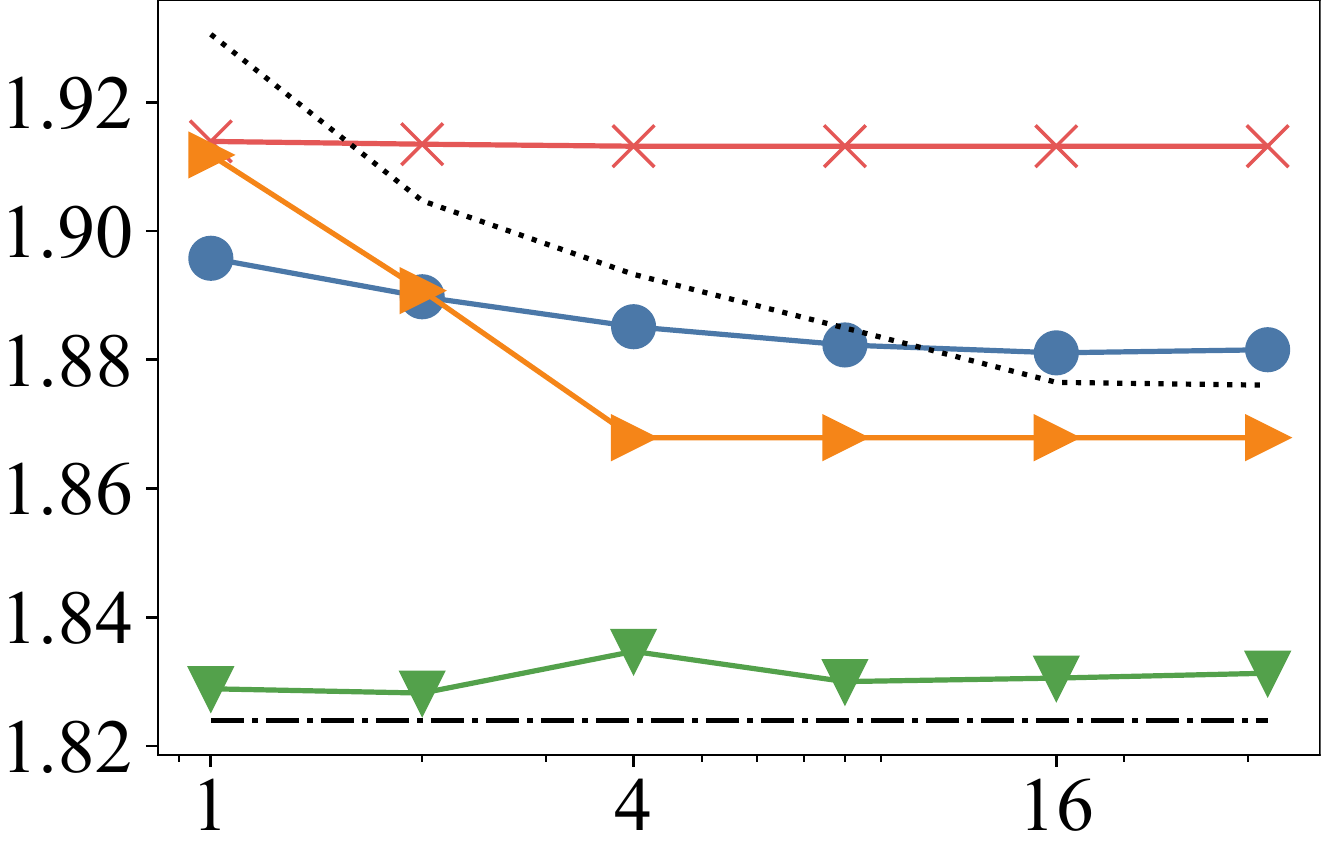}
            &
            \rotatebox[origin=lt]{90}{\hspace{\ndcghspace} \small nDCG@10}
            &
            \includegraphics[scale=\performancescale]{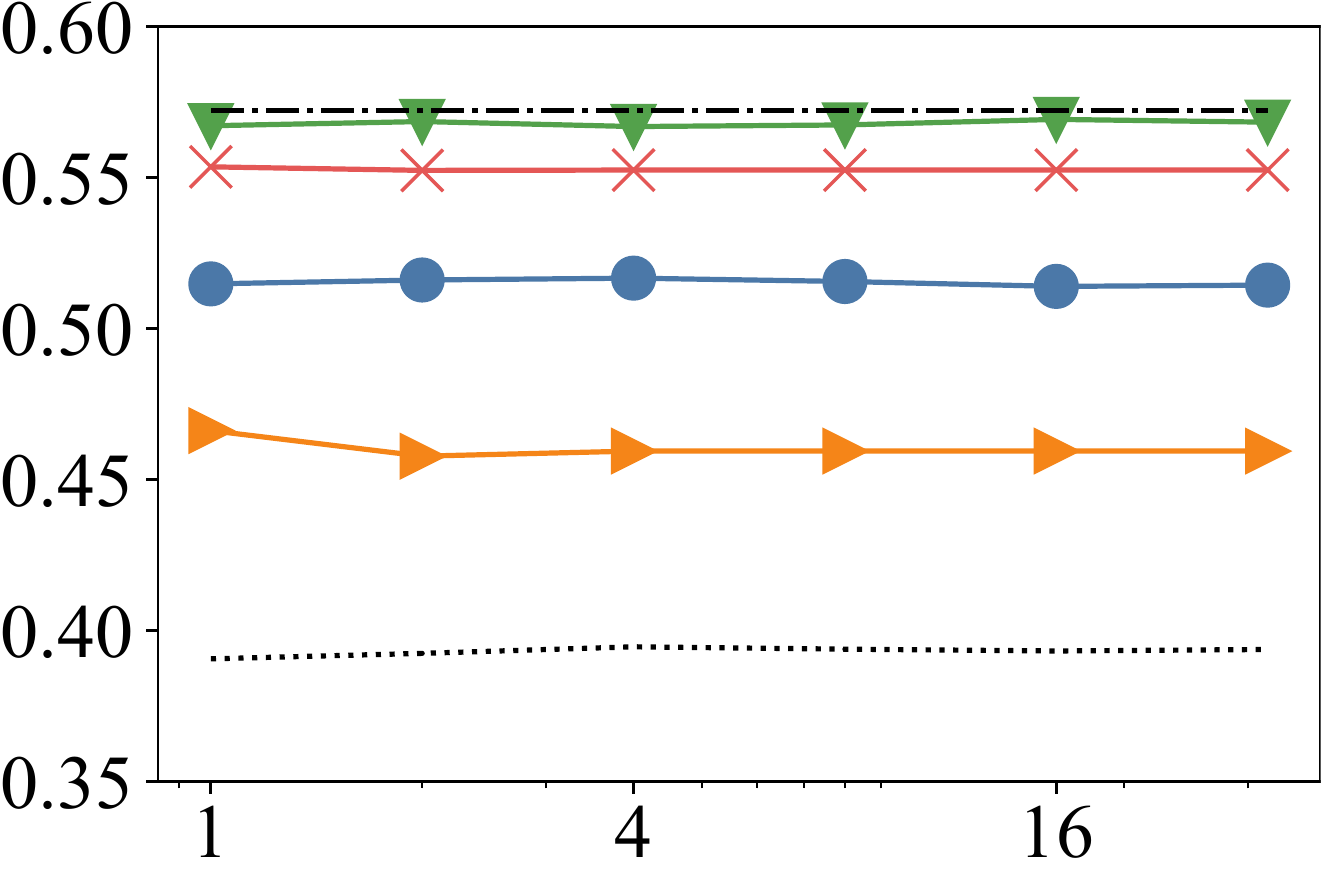} 
            & &
            \rotatebox[origin=lt]{90}{\hspace{\fairnesshspace} \small EEL}
            &
            \includegraphics[scale=\performancescale]{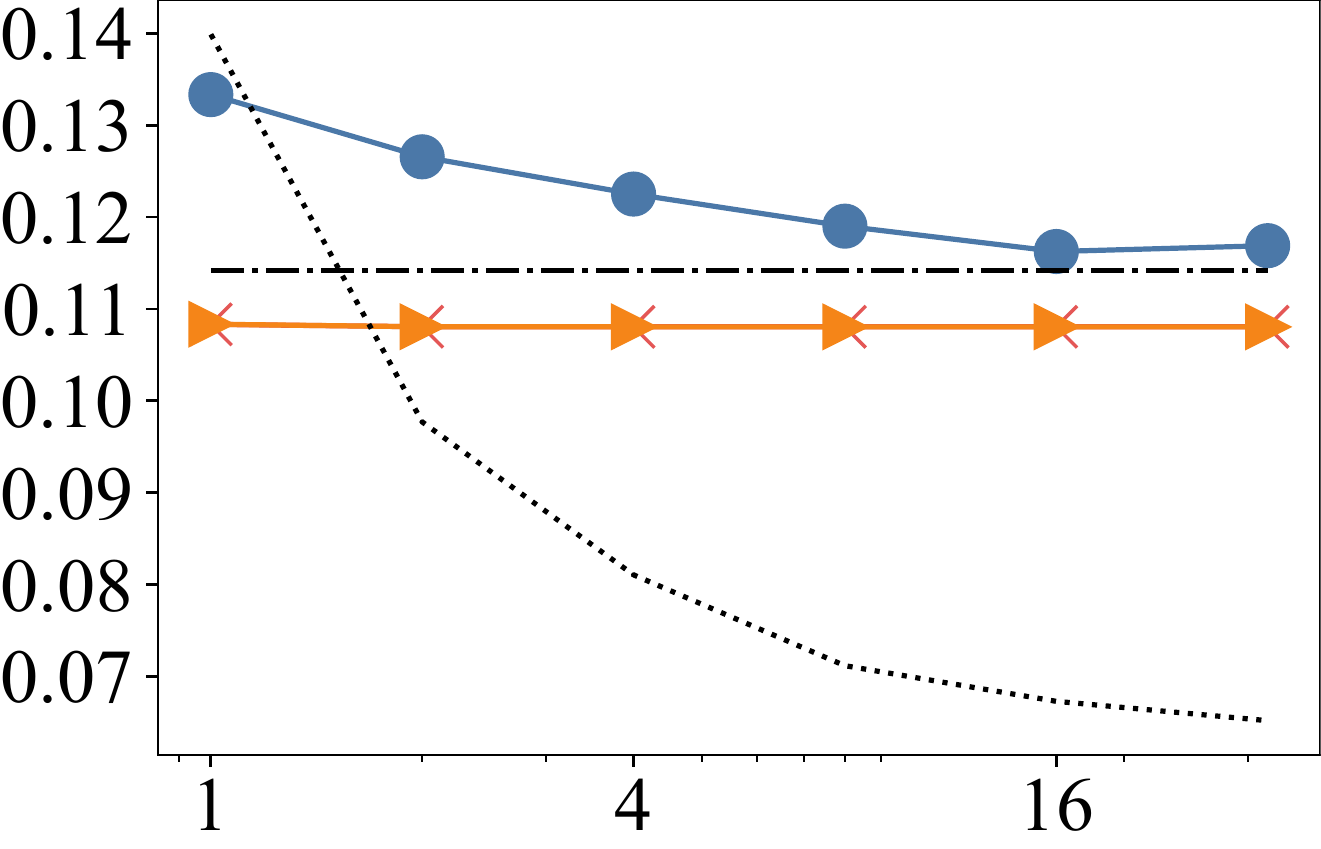}
            &
            \rotatebox[origin=lt]{90}{\hspace{\ndcghspace} \small nDCG@10}
            &
            \includegraphics[scale=\performancescale]{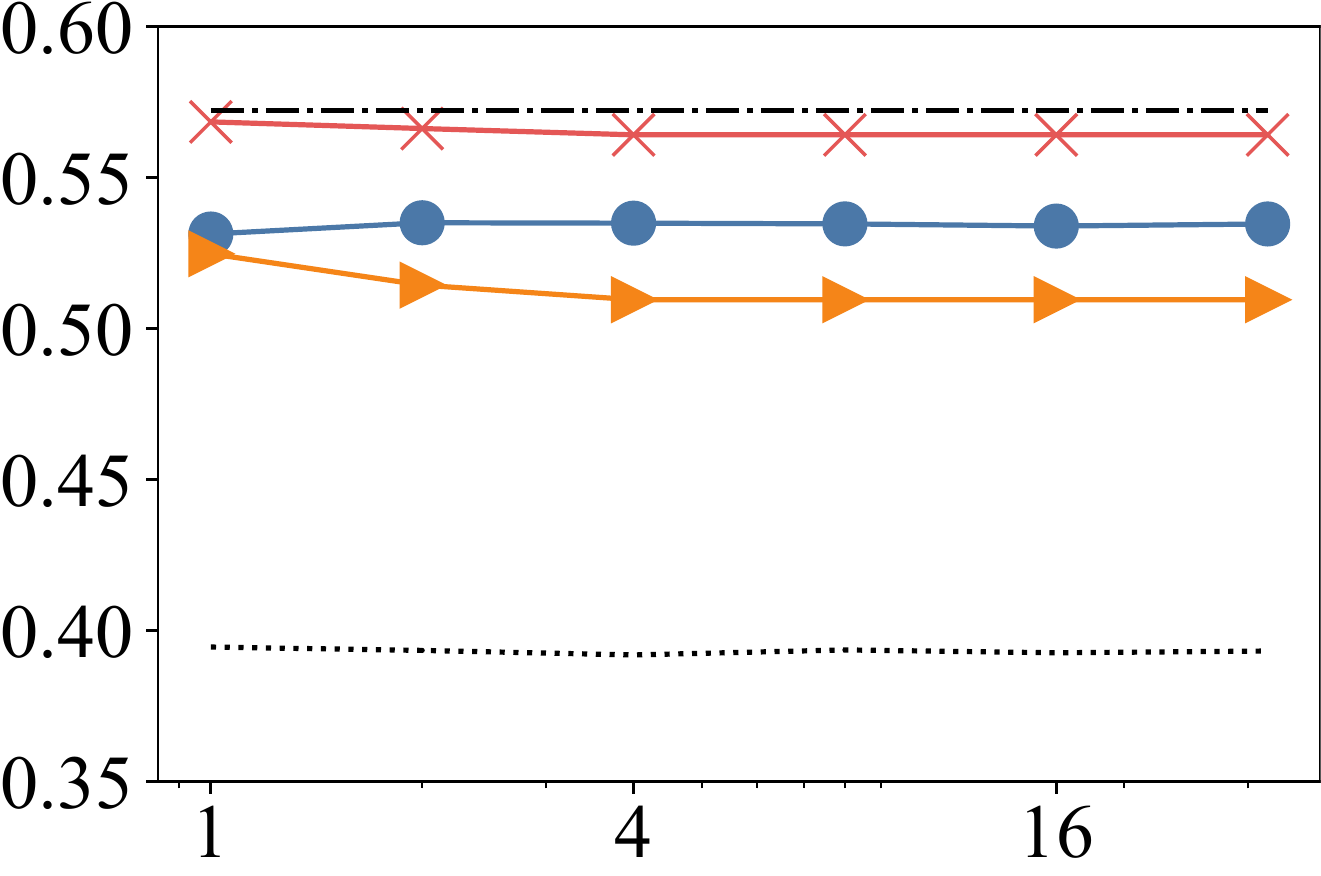} \\
            & \small Sessions &
            & \small Sessions &
            &
            & \small Sessions &
            & \small Sessions \\
            \multicolumn{9}{c}{\includegraphics[scale=0.4]{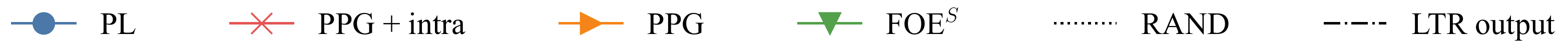}}
            
        \end{tabular}
        \caption{\emph{LTR outputs as utility estimates.} Performance comparison of PL and \ourmethod for optimizing different fairness metrics: DTR and EEL (lower is better in both metrics).
        Top: TREC 2019; middle: TREC 2020; bottom: MSLR. In all cases \ourmethod achieves comparable results to the baselines (see Appendix~\ref{sec:table} for more details).}
        \label{fig:performance}
    \end{figure*}
    }

%% file: sections/figure_tex/05-fig-performance-true.tex
{
    \setlength{\tabcolsep}{1pt}
    \newcommand{\performancescale}{0.28}
    \newcommand{\fairnesshspace}{2.5em}
    \newcommand{\ndcghspace}{1em}

\begin{figure*}[t]
    \centering
    \begin{tabular}[]{rcrccrcrc}

        \rotatebox[origin=lt]{90}{\hspace{\fairnesshspace} \small DTR}
        &
        \includegraphics[scale=\performancescale]{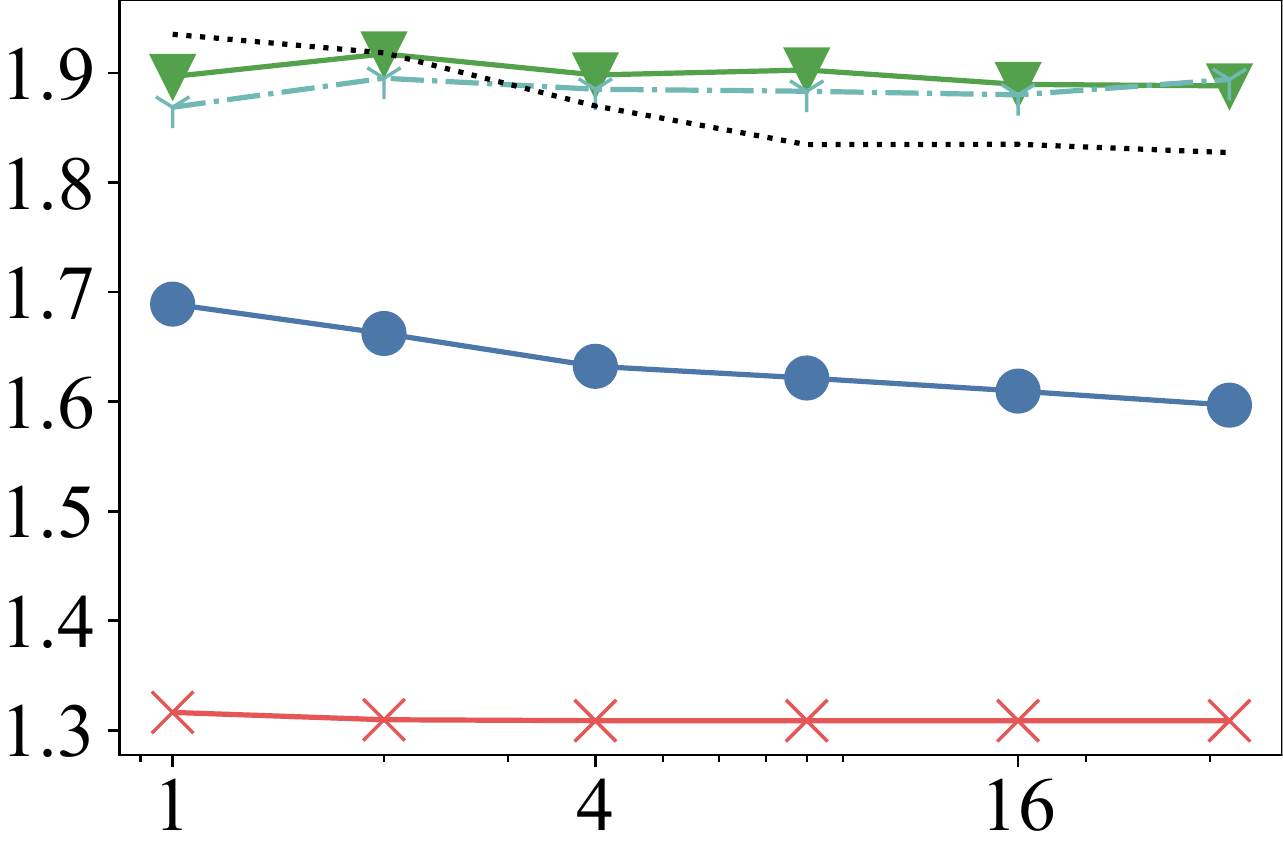}
        &
        \rotatebox[origin=lt]{90}{\hspace{\ndcghspace} \small nDCG@10}
        &
        \includegraphics[scale=\performancescale]{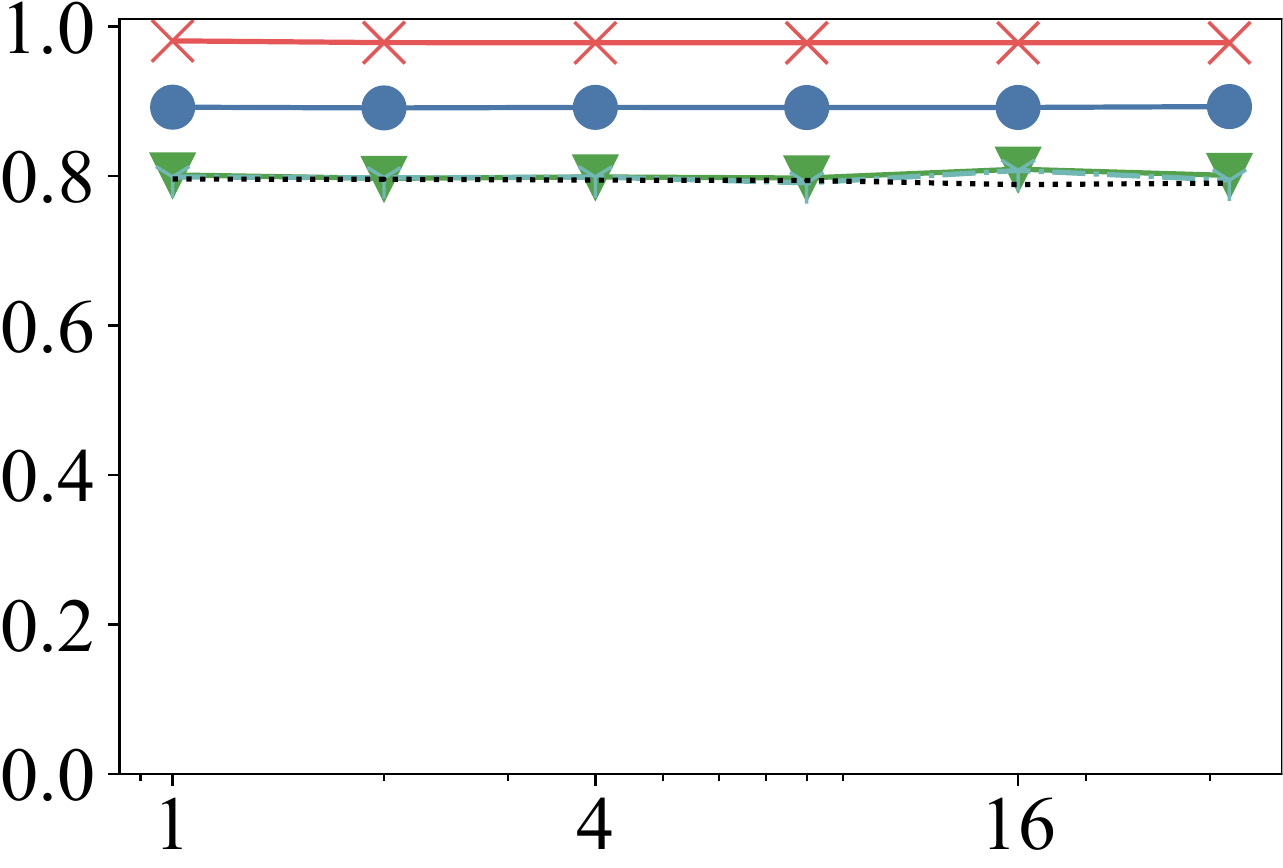} 
        & &
        \rotatebox[origin=lt]{90}{\hspace{\fairnesshspace} \small EEL}
        &
        \includegraphics[scale=\performancescale]{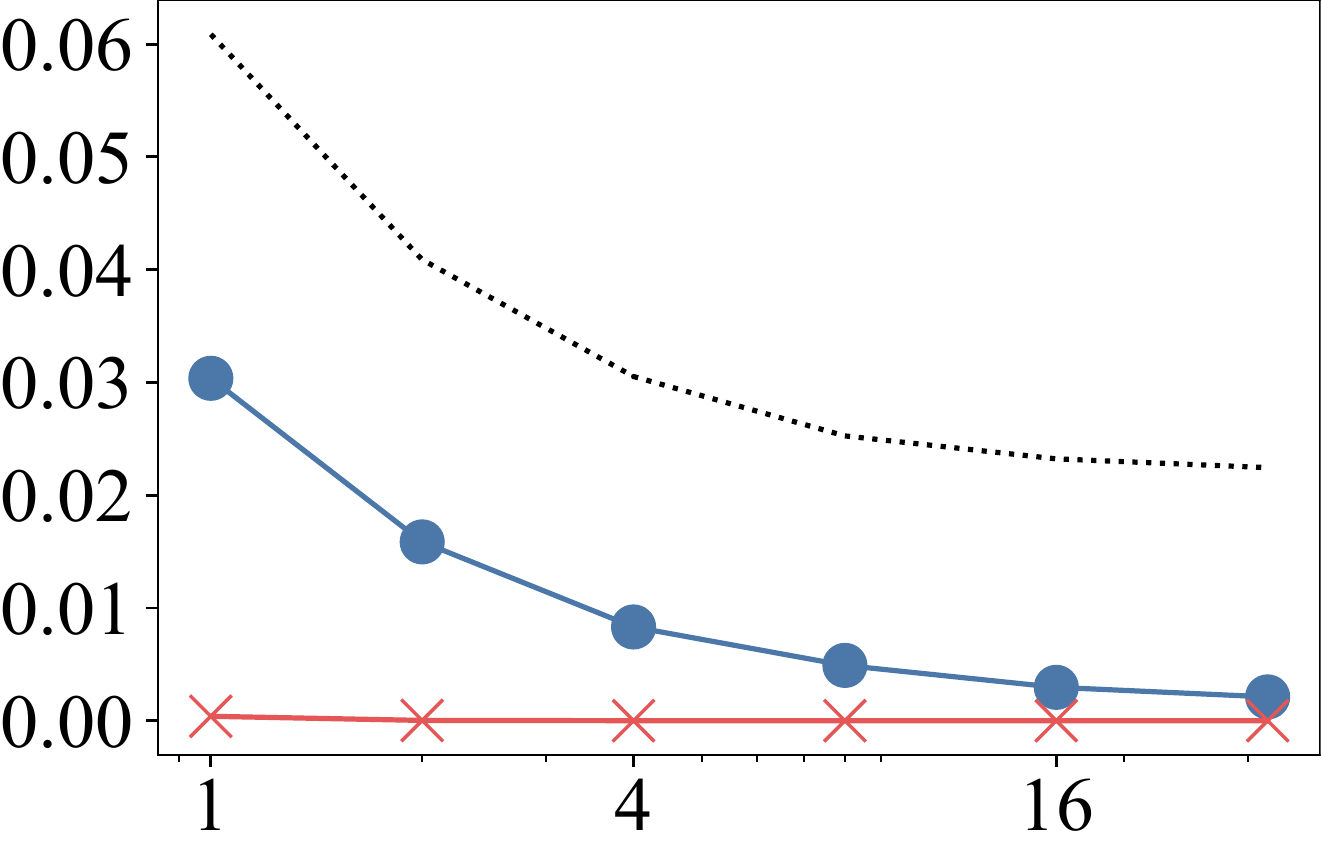}
        &
        \rotatebox[origin=lt]{90}{\hspace{\ndcghspace} \small nDCG@10}
        &
        \includegraphics[scale=\performancescale]{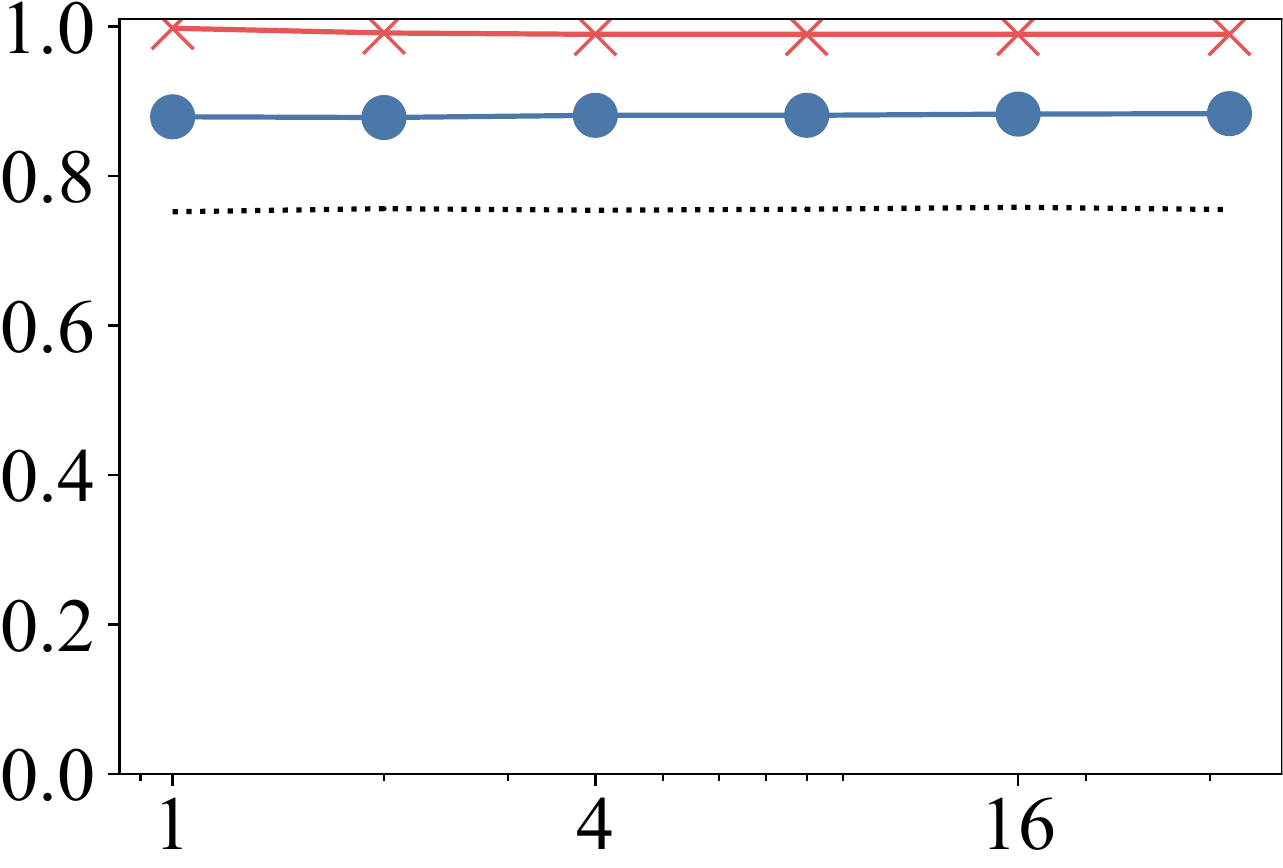} \\

        \rotatebox[origin=lt]{90}{\hspace{\fairnesshspace} \small DTR}
        &
        \includegraphics[scale=\performancescale]{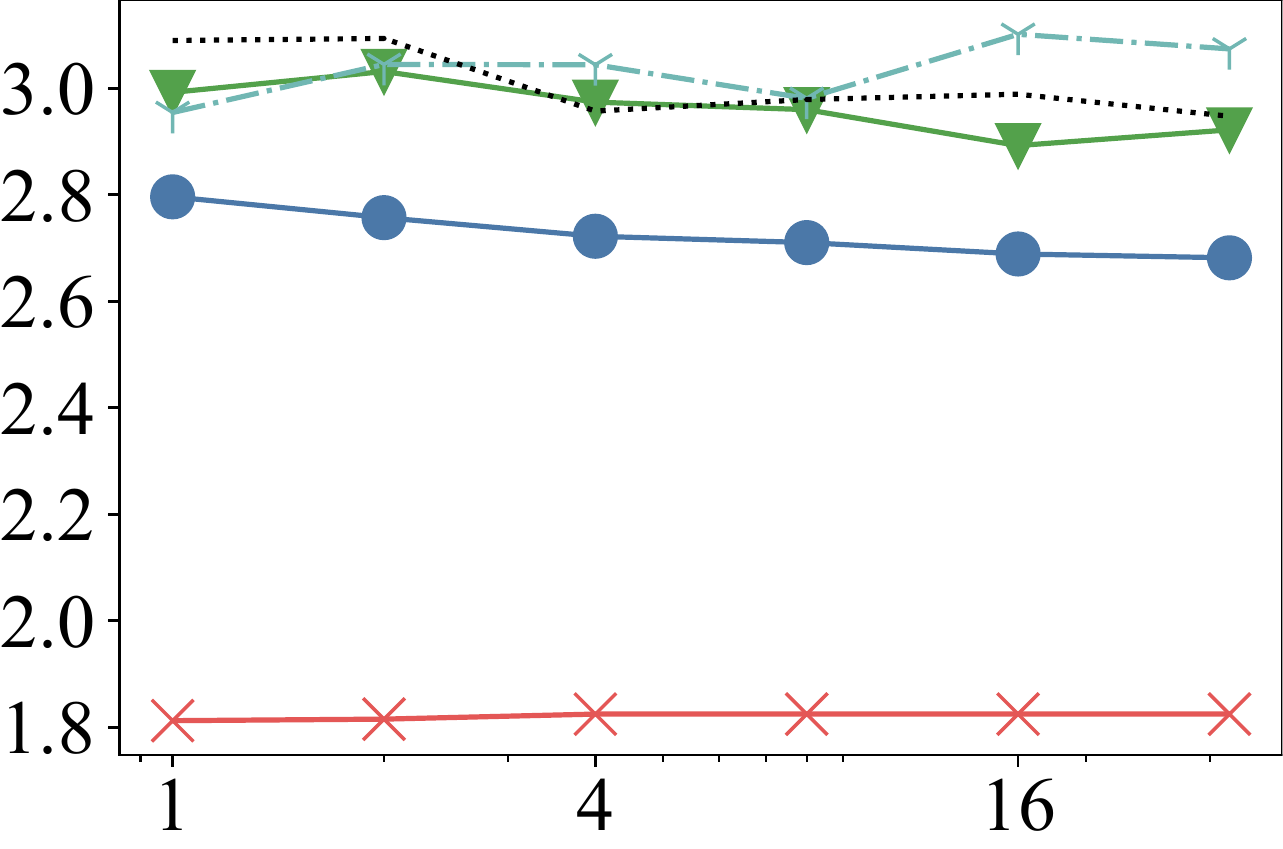}
        &
        \rotatebox[origin=lt]{90}{\hspace{\ndcghspace} \small nDCG@10}
        &
        \includegraphics[scale=\performancescale]{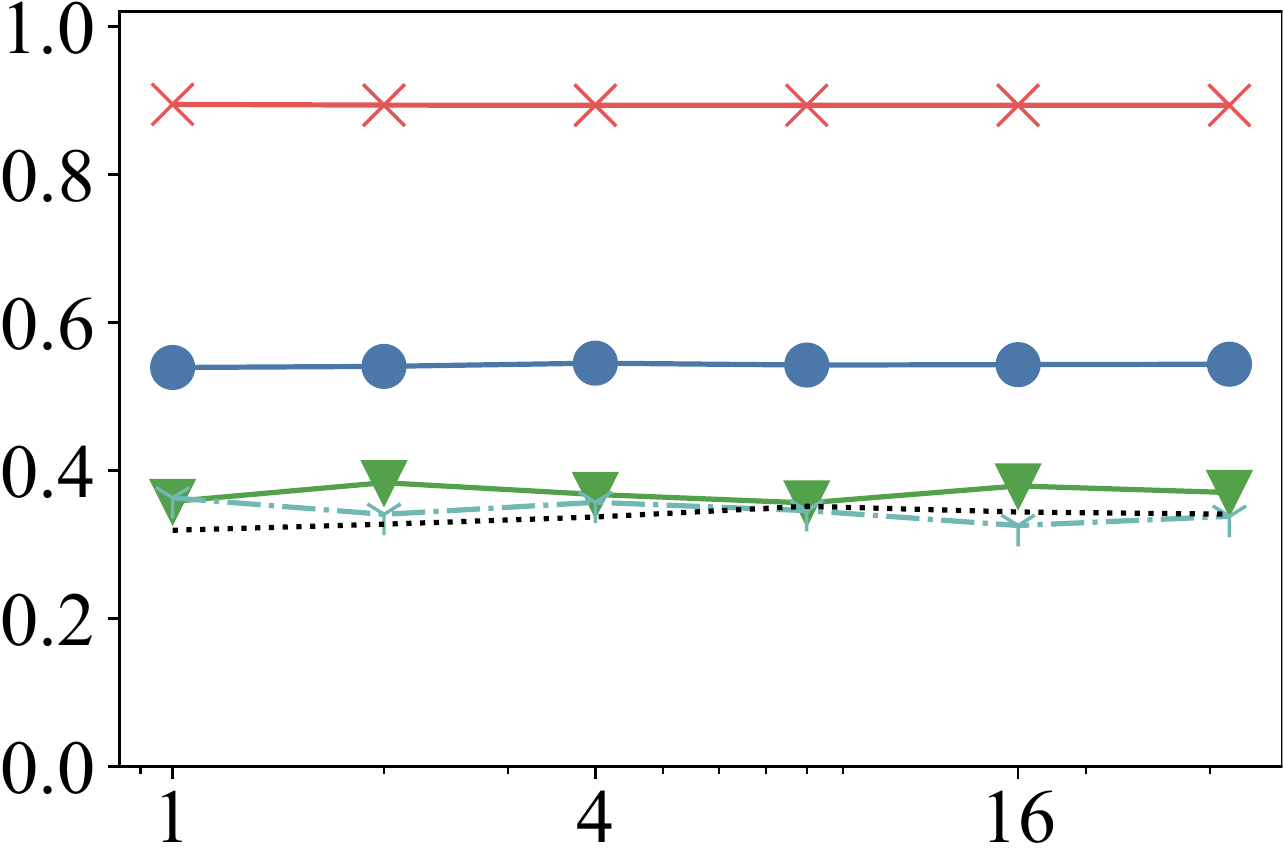} 
        & &
        \rotatebox[origin=lt]{90}{\hspace{\fairnesshspace} \small EEL}
        &
        \includegraphics[scale=\performancescale]{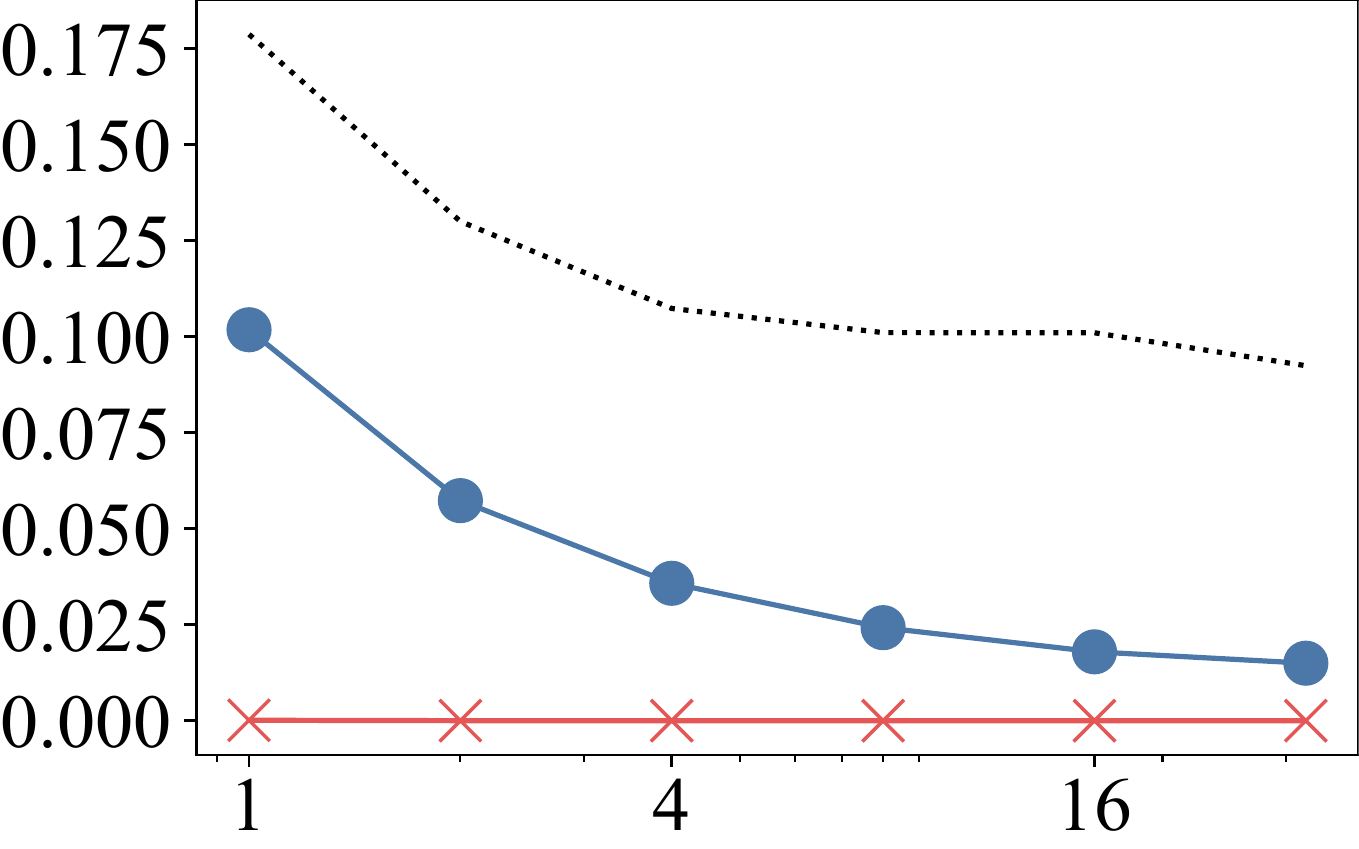}
        &
        \rotatebox[origin=lt]{90}{\hspace{\ndcghspace} \small nDCG@10}
        &
        \includegraphics[scale=\performancescale]{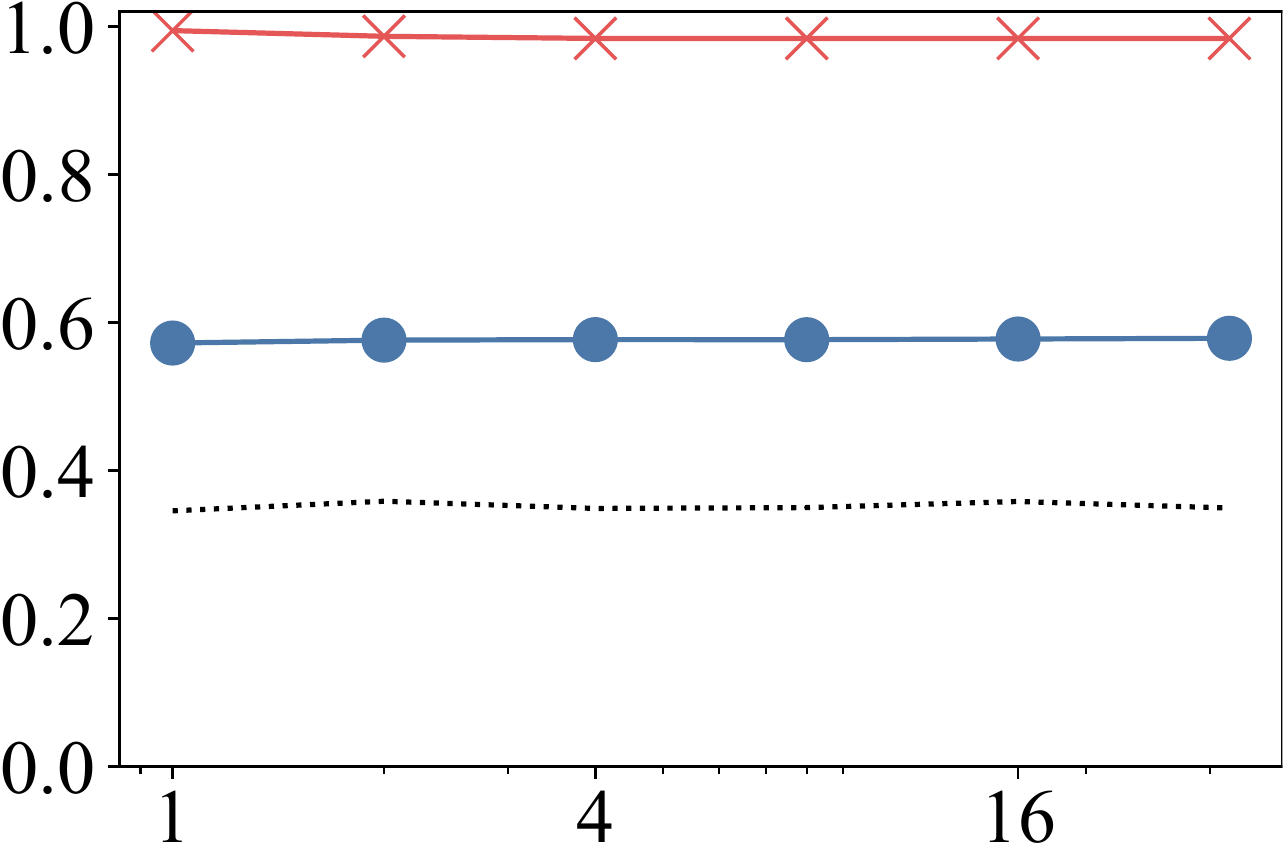} \\

        \rotatebox[origin=lt]{90}{\hspace{\fairnesshspace} \small DTR}
        &
        \includegraphics[scale=\performancescale]{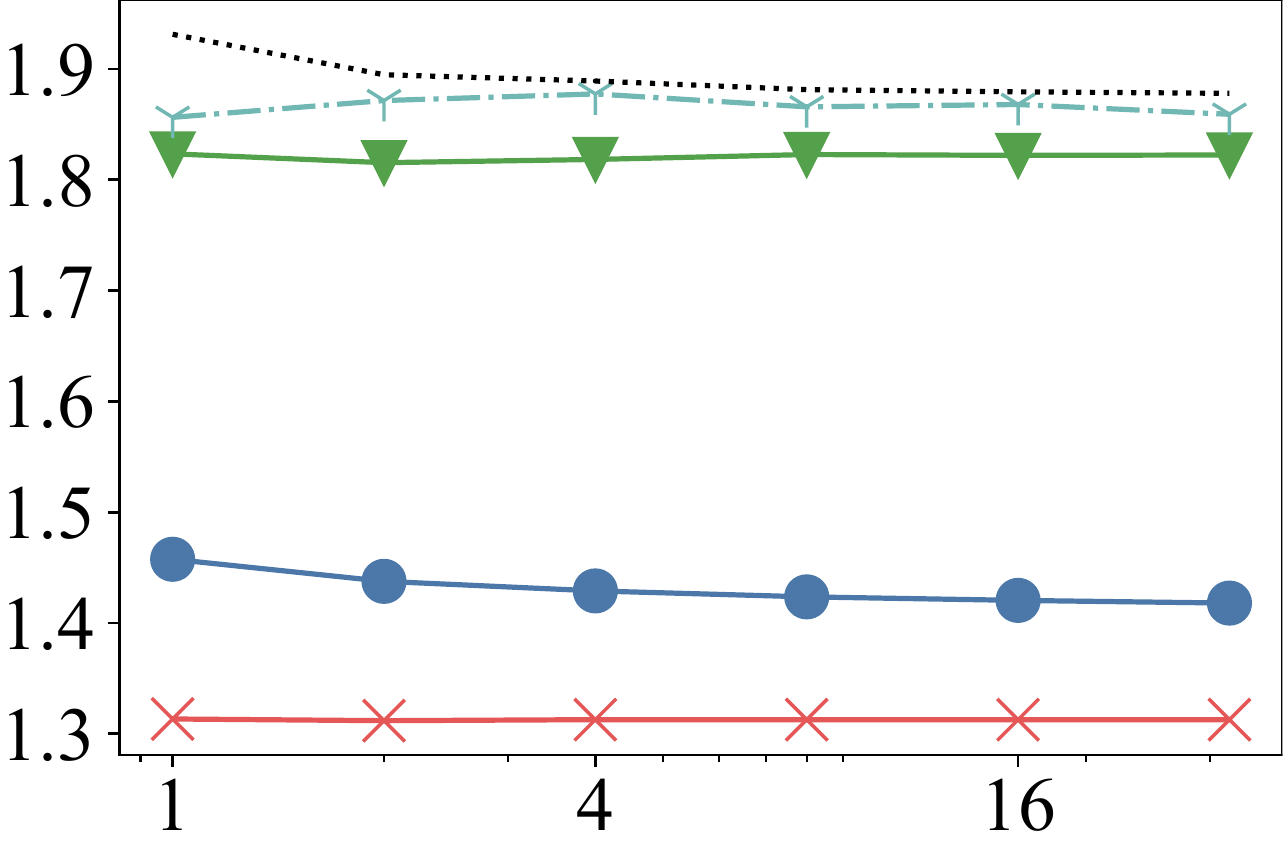}
        &
        \rotatebox[origin=lt]{90}{\hspace{\ndcghspace} \small nDCG@10}
        &
        \includegraphics[scale=\performancescale]{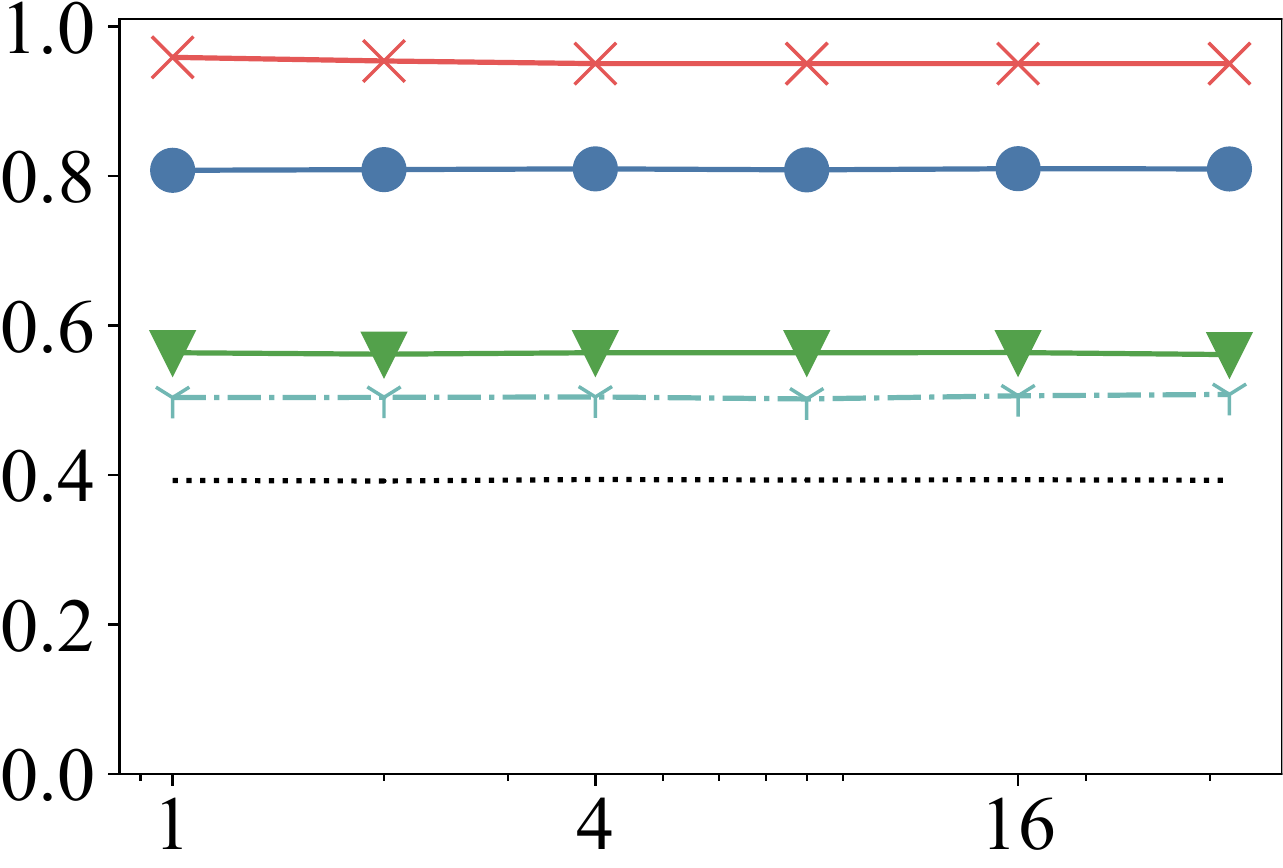} 
        & &
        \rotatebox[origin=lt]{90}{\hspace{\fairnesshspace} \small EEL}
        &
        \includegraphics[scale=\performancescale]{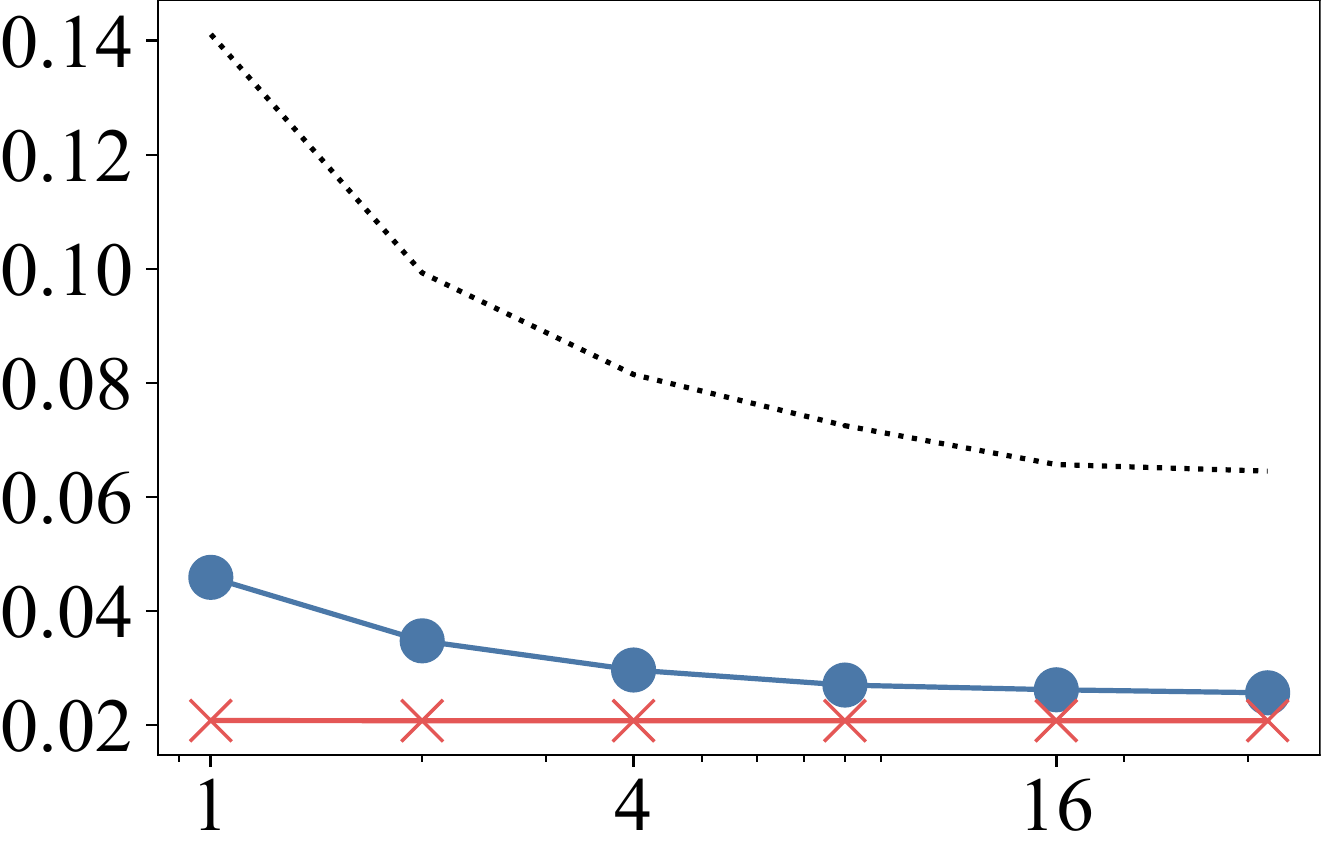}
        &
        \rotatebox[origin=lt]{90}{\hspace{\ndcghspace} \small nDCG@10}
        &
        \includegraphics[scale=\performancescale]{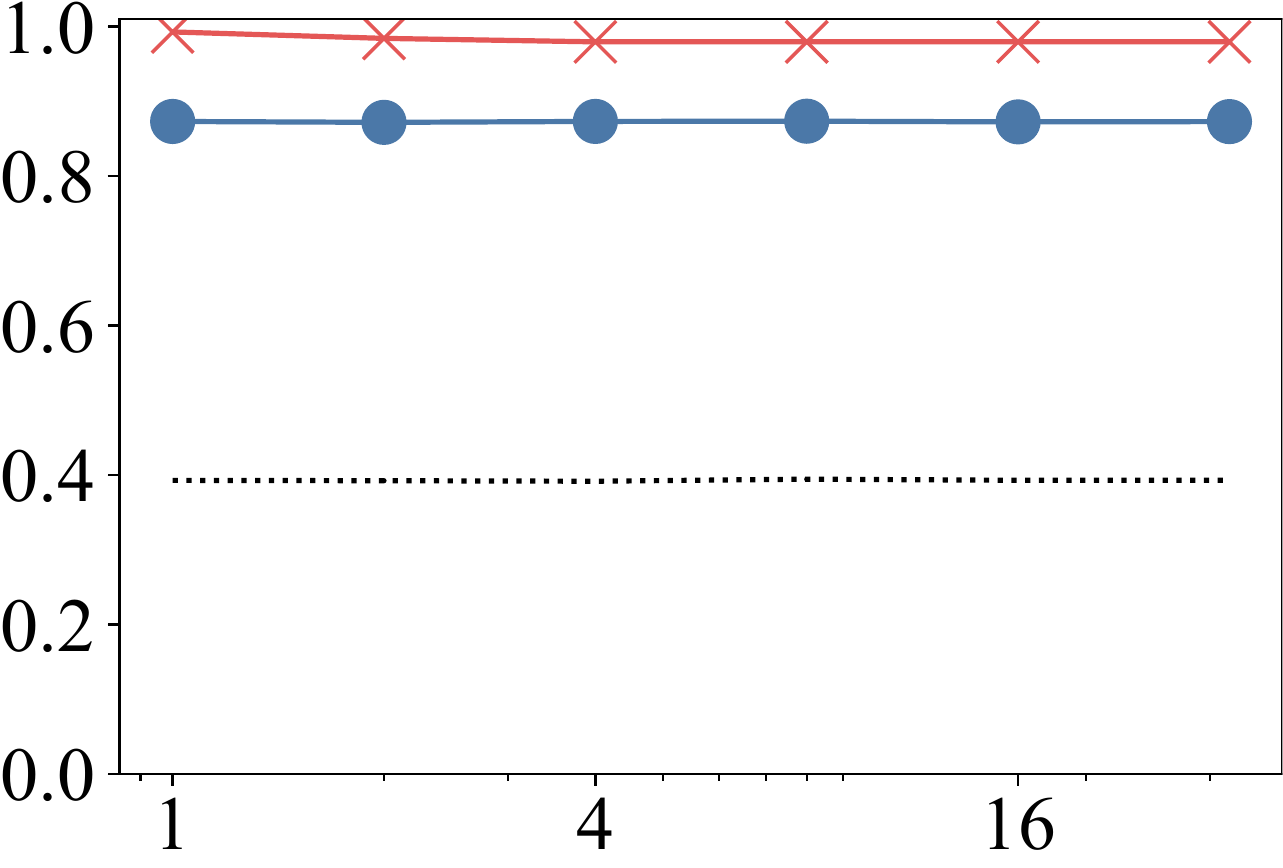} \\
        & \small Sessions &
        & \small Sessions &
        &
        & \small Sessions &
        & \small Sessions \\
        \multicolumn{9}{c}{\includegraphics[scale=0.4]{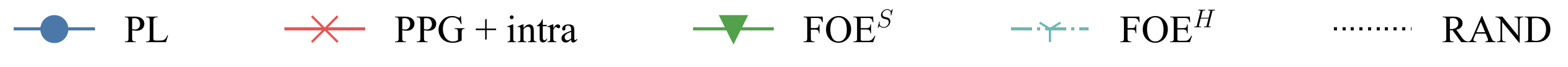}}
        
    \end{tabular}
    \caption{\emph{True relevance labels as utility estimates.} Performance comparison of PL and \ourmethod for optimizing different fairness metrics: DTR and EEL.
    Top: TREC 2019; middle: TREC 2020; bottom: MSLR (see Appendix~\ref{sec:table} for more details).}
    \label{fig:performanceknown}
\end{figure*}
}

%% file: sections/06-conclusion.tex

\section{Conclusion}
\label{sec:conclusion} 
In this paper, we have defined~\acf{PPG}, a novel representation for permutation distributions.
We used PPG as a substitute for PL in black-box optimization of fairness metrics, using the REINFORCE algorithm.
Unlike PL which is represented by pointwise logits, PPG is constructed by a reference permutation together with pairwise inversion probabilities.
The reference permutation of PPG is very useful in deterministic scenarios.

Our experiments show that \ourmethod, compared to state-of-the-art post-processing fairness optimization methods, is more robust in finding a deterministic fair permutation for one session, while having comparable performance for expected fairness over larger numbers of sessions.
The pairwise inversion probabilities allow us to impose pairwise constraints that can control other objectives while optimizing for fairness.
We experimentally verified the effectiveness of such pairwise constraints on controlling the ranking performance.
Finally, we have shown that in scenarios such as tabular search, where high quality estimates of utility are available, PPG performs outstandingly well, with a considerable gap to other state-of-the-art fairness optimization methods.

As future work, it would be interesting to test the effectiveness of PPG on other fairness metrics as well as other general permutation optimization problems.
Another possible line of work is to apply PPG for an in-processing method as a substitute for PL.

%% file: sections/0A-appendix.tex
\section{Log Derivative of Probability Distribution}
\label{sec:logderivativapprox}
Here we discuss the approximate log derivative of the probability distribution, given in Eq.~\eqref{eq:ppggrad}.
Rewriting Eq.~\eqref{eq:ppggradfull} we have:
\begin{equation*}
        \frac{\partial \log P(E_\pi)}{\partial w_e} = \underbrace{\frac{\mathbb{I}\bigl(e\in E_\pi\bigr) - w_e}{w_e \cdot \bigl(1 - w_e\bigr)}}_{\alpha_w}
        - \underbrace{\frac{1}{\permcond}\frac{\partial \permcond}{\partial w_e}}_{\beta_{\rho,w}}.
\end{equation*}
\noindent
The first term, $\alpha_w$, only depends on the weight $w_e$ and sampled graph $E_\pi$.
However, the second term, $\beta_{\rho, w}$, depends on $\rho$ which cannot be calculated feasibly because of the exponential size of the permutation space.
Consequently, we explain why it is save to consider $\alpha_w-\beta_{\rho, w} \approx \alpha_w$ in gradient descent.
First note that, fixing all the weights other than $w_e$, $\permcond$ is a linear function of $w_e$:
\begin{equation*}
    \permcond = w_e\cdot A + \bigl(1 - w_e\bigr) \cdot B,
\end{equation*} 
\noindent
where $A$ is the probability sum of all the permutation graphs containing $e$ and $B$ is the probability sum of all the permutation graphs not containing $e$.
Therefore, we have
\begin{equation*}
    \beta_{\rho, w} = \frac{A-B}{w_e\cdot\bigl(A-B\bigr) + B}.
\end{equation*}
We know from graph theory that~\citep{dushnik1941partially}:
\begin{equation}
    \label{eq:complementgraph}
    E \in \permspace \Rightarrow \bar{E} \in \permspace,
\end{equation}
\noindent
where $\bar{E}$ is the complement graph of $E$.
For training the PPG weights, we initialize all the weights to $0.5$ and slightly change them using the gradients.
Simple counting shows that, when all the edges have a $0.5$ probability of sampling, we have:
\begin{equation}
    \label{eq:counting}
    A = B = \frac{n!}{2^{\frac{n(n-1)}{2}}},
\end{equation}
\noindent
which means that for the initial weights $\beta_{\rho, w} = 0$.
We further show that, even when $\beta_{\rho, w} \neq 0$, the gradient and $\alpha_w$ \emph{always} have the same sign.
This is a critical condition in gradient descent with small learning rate, as the weights are guaranteed to update in the correct \emph{direction}.
To show $\alpha_w$ and $\alpha_w - \beta_{\rho, w}$ always have the same sign, we notice that when $\alpha_w$ and $\beta_{\rho, w}$ have different signs, the proposition is true.
It remains to prove the proposition for the case where $\alpha_w$ and $\beta_{\rho, w}$ have the same signs:
\begin{enumerate}[leftmargin=*]
    \item {Case 1:} $\alpha_w = \frac{1}{w_e} > 0$ and $\beta_{\rho, w} > 0$
    \begin{equation}
        w_e(A-B) + B > A - B \xRightarrow{(A-B > 0)} \beta_{\rho, w} < \frac{1}{w_e}.
    \end{equation}
    \item {Case 2:} $\alpha_w = \frac{-1}{1 - w_e} < 0$ and $\beta_{\rho, w} < 0$
    \begin{equation}
        \begin{split}
        & A\cdot w_e + B(1-w_e) > -(A-B)(1-w_e)>0 \\
        & \xRightarrow{(A-B < 0)} -\beta_{\rho, w} < \frac{1}{1-w_e}.
        \end{split}
    \end{equation}
\end{enumerate}
\noindent
To sum up, we have shown that in the working range of weights, i.e. $\approx 0.5$, we have $\beta_{\rho,w}\approx 0$.
And more importantly, in general, $\alpha_w$ and $\alpha_w-\beta_{\rho,w}$ always have the same sign, ensuring that the gradient descent updates the weights in the correct direction.
Our experiments show a very negligible sensitivity of PPG search to the learning rate, which translates to negligible sensitivity to the gradient size.\looseness=-1

%% file: sections/0B-table.tex
\section{Results for Small Number of Sessions}
\label{sec:table}
\input{sections/figure_tex/05-tbl-performance}

Table~\ref{tbl:resultsltr} and~\ref{tbl:resultstruerel} contain the results of all models on 4 sessions with LTR outputs as utility estimates and knowledge of the true relevance labels, respectively. 
In the case of a limited number of sessions and LTR outputs as utility estimates (Table~\ref{tbl:resultsltr}), PPG has a slight advantage compared to PL when optimizing for both EEL and DTR in all three tested datasets. 
In this case, FOE$^S$, specifically designed for DTR optimization, leads to the fairest ranking in all three datasets.
When the true relevance labels are known (Table~\ref{tbl:resultstruerel}),~\ourmethod outperforms the baselines by a noticeable margin (20\%, 33\%, and 8\% relative decrease for DTR and 100\%, 100\%, and 30\% relative decrease for EEL, compared to the runner up, in TREC 2019, TREC 2020, and MSLR, respectively), obtaining the fairest ranking policies while achieving NDCG scores close to the ideal ranking.
Note that the scores reported in Table~\ref{tbl:resultsltr} and~\ref{tbl:resultstruerel} are from two separate experiments which explains the small differences in the results for the RAND baseline.

\input{sections/figure_tex/05-tbl-performance-true}

%% file: sections/figure_tex/05-tbl-performance.tex
\begin{table}[t]
  \caption{Results on 4 sessions with LTR generated labels.}
  \label{tbl:resultsltr}
  \setlength{\tabcolsep}{12pt}
  \centering
  \resizebox{1.0\columnwidth}{!}{%
  \begin{tabular}{l@{}clcccc}
    \toprule
    \multicolumn{2}{c}{} &
    \multicolumn{1}{c}{} &
    \multicolumn{2}{c}{Fairness } &
    \multicolumn{2}{c}{NDCG@10 } 
      \\
    \cmidrule(r){4-5}
    \cmidrule(r){6-7}
  
  &
      &   
    \multicolumn{1}{c}{Model} &
    \multicolumn{1}{c}{DTR{\tiny ~$\downarrow$}} &
    \multicolumn{1}{c}{EEL{\tiny ~$\downarrow$}} &
    \multicolumn{1}{c}{DTR{\tiny ~$\uparrow$}} &
    \multicolumn{1}{c}{EEL{\tiny ~$\uparrow$}} \\
    \midrule
      \multirow{8}{*}{\rotatebox[origin=c]{90}{\textbf{  TREC 2019}}}  
      & & $FOE^S$&{\bfseries 1.898}&-&{0.805}&-\\
      & & $FOE^H$&{1.944}&-&{0.803}&-\\
      & & PL&{1.939}&{0.052}&{\bfseries 0.817}&{\bfseries 0.793}\\
      & & PPG&{1.925}&{\bfseries 0.047}&{0.806}&{0.780}\\
      & & PPG + intra&{1.938}&{\bfseries 0.047}&{0.812}&{0.790}\\
      \cmidrule{3-7}
      & & LTR output&{1.901}&{0.053}&{0.816}&{0.798}\\
      & & RAND&{1.848}&{0.030}&{0.792}&{0.754}\\
      \midrule 
        
        \multirow{8}{*}{\rotatebox[origin=c]{90}{\textbf{TREC 2020}}}  
        & & $FOE^S$&{\bfseries 2.966}&-&{\bfseries 0.376}&-\\
        & & $FOE^H$&{3.047}&-&{0.371}&-\\
        & & PL&{2.981}&{0.180}&{0.369}&{\bfseries 0.388}\\
        & & PPG&{2.969}&{\bfseries 0.158}&{0.361}&{0.378}\\
        & & PPG + intra&{2.979}&{\bfseries 0.158}&{0.370}&{0.379}\\
        \cmidrule{3-7}
        & & LTR output&{3.004}&{0.173}&{0.373}&{0.386}\\
        & & RAND&{2.948}&{0.118}&{0.334}&{0.352}\\
        \midrule 
        
        \multirow{8}{*}{\rotatebox[origin=c]{90}{\textbf{MSLR-qs}}}  
        & & $FOE^S$&{\bfseries 1.835}&-&{\bfseries 0.567}&-\\
        & & $FOE^H$&{1.842}&-&{0.526}&-\\
        & & PL&{1.885}&{0.123}&{0.517}&{0.535}\\
        & & PPG&{1.868}&{\bfseries 0.108}&{0.459}&{0.510}\\
        & & PPG + intra&{1.913}&{\bfseries 0.108}&{0.552}&{\bfseries 0.564}\\
        \cmidrule{3-7}
        & & LTR output&{1.824}&{0.114}&{0.572}&{0.572}\\
        & & RAND&{1.893}&{0.081}&{0.395}&{0.392}\\
    \bottomrule
          
  \end{tabular}%
  }
\end{table}

%% file: sections/figure_tex/05-tbl-performance-true.tex
\begin{table}[!h]
  \caption{Results on 4 sessions with true relevance labels.}
  \label{tbl:resultstruerel}
  \setlength{\tabcolsep}{12pt}
  \centering
  \resizebox{1.0\columnwidth}{!}{%
  \begin{tabular}{l@{}clcccc}
    \toprule
    \multicolumn{2}{c}{} &
    \multicolumn{1}{c}{} &
    \multicolumn{2}{c}{Fairness } &
    \multicolumn{2}{c}{NDCG@10 } 
      \\
    \cmidrule(r){4-5}
    \cmidrule(r){6-7}
  
  &
      &   
    \multicolumn{1}{c}{Model} &
    \multicolumn{1}{c}{DTR{\tiny ~$\downarrow$}} &
    \multicolumn{1}{c}{EEL{\tiny ~$\downarrow$}} &
    \multicolumn{1}{c}{DTR{\tiny ~$\uparrow$}} &
    \multicolumn{1}{c}{EEL{\tiny ~$\uparrow$}} \\
    \midrule
      \multirow{8}{*}{\rotatebox[origin=c]{90}{\textbf{  TREC 2019}}}  
      & & $FOE^S$&{1.898}&-&{0.799}&-\\
      & & $FOE^H$&{1.885}&-&{0.799}&-\\
      & & PL&{1.632}&{0.008}&{0.892}&{0.881}\\
      & & PPG&{1.320}&{\bfseries 0.000}&{0.881}&{0.900}\\
      & & PPG + intra&{\bfseries 1.309}&{\bfseries 0.000}&{\bfseries 0.978}&{\bfseries 0.989}\\
      \cmidrule{3-7}
      & & Ideal ranking&{1.668}&{0.014}&{1.000}&{1.000}\\
      & & RAND&{1.870}&{0.031}&{0.794}&{0.754}\\
      \midrule 
        
        \multirow{8}{*}{\rotatebox[origin=c]{90}{\textbf{TREC 2020}}}  
        & & $FOE^S$&{2.974}&-&{0.368}&-\\
        & & $FOE^H$&{3.044}&-&{0.357}&-\\
        & & PL&{2.722}&{0.036}&{0.545}&{0.577}\\
        & & PPG&{2.141}&{\bfseries 0.000}&{0.574}&{0.819}\\
        & & PPG + intra&{\bfseries 1.825}&{\bfseries 0.000}&{\bfseries 0.894}&{\bfseries 0.984}\\
        \cmidrule{3-7}
        & & Ideal ranking&{2.538}&{0.023}&{1.000}&{1.000}\\
        & & RAND&{2.957}&{0.107}&{0.337}&{0.349}\\
        \midrule 
        
        \multirow{8}{*}{\rotatebox[origin=c]{90}{\textbf{MSLR-qs}}}  
        & & $FOE^S$&{1.818}&-&{0.564}&-\\
        & & $FOE^H$&{1.877}&-&{0.504}&-\\
        & & PL&{1.429}&{0.030}&{0.810}&{0.873}\\
        & & PPG&{1.395}&{\bfseries 0.021}&{0.619}&{0.821}\\
        & & PPG + intra&{\bfseries 1.313}&{\bfseries 0.021}&{\bfseries 0.950}&{\bfseries 0.979}\\
        \cmidrule{3-7}
        & & Ideal ranking&{1.548}&{0.027}&{1.000}&{1.000}\\
        & & RAND&{1.889}&{0.081}&{0.394}&{0.392}\\
    \bottomrule
          
  \end{tabular}%
  }
\end{table}